\documentclass{article} %
\usepackage{iclr2027_conference,times}

\usepackage{amsmath,amsfonts,bm}

\def\eqref#1{equation~\ref{#1}}
\def\1{\bm{1}}

\DeclareMathAlphabet{\mathsfit}{\encodingdefault}{\sfdefault}{m}{sl}
\SetMathAlphabet{\mathsfit}{bold}{\encodingdefault}{\sfdefault}{bx}{n}

\usepackage{hyperref}
\usepackage{url}
\usepackage{enumitem}

\title{Multi-Agent Egocentric World Model\\ with Fine-Grained Embodied Interaction}

\author{%
\textbf{Dahyun Chung} \quad
\textbf{Siyoon Jin} \quad
\textbf{Hyunwook Choi} \quad
\textbf{Honggyu An} \quad
\textbf{Junyoung Seo} \\
\textbf{Hyunsung Kim} \quad
\textbf{Seung Wook Kim}$^\dagger$ \quad
\textbf{Seungryong Kim}$^\dagger$\\
[4pt]
$^\dagger$ Co-corresponding authors\\
[4pt]
KAIST AI\\
[3pt]
\href{https://cvlab-kaist.github.io/ME-World}{\texttt{https://cvlab-kaist.github.io/ME-World}}
}

\newcommand{\paragrapht}[1]{\noindent\textbf{#1}}
\usepackage{graphicx}
\usepackage{booktabs}
\usepackage{wrapfig}
\usepackage{pifont}
\usepackage{bbm}
\usepackage{xspace}

\usepackage{multirow}
\usepackage{multicol}
\usepackage{adjustbox}
\usepackage[table]{xcolor}
\definecolor{bestgreen}{HTML}{A8D8F0}
\definecolor{secondgreen}{HTML}{E0F2FA}
\newcommand{\ours}{ME-World\xspace}
\newcommand{\best}[1]{\cellcolor{bestgreen}\textbf{#1}}
\newcommand{\second}[1]{\cellcolor{secondgreen}\underline{#1}}

\usepackage{lipsum}

\begin{document}
\maketitle

\begin{figure}[htbp]
    \centering
    \includegraphics[width=\linewidth]{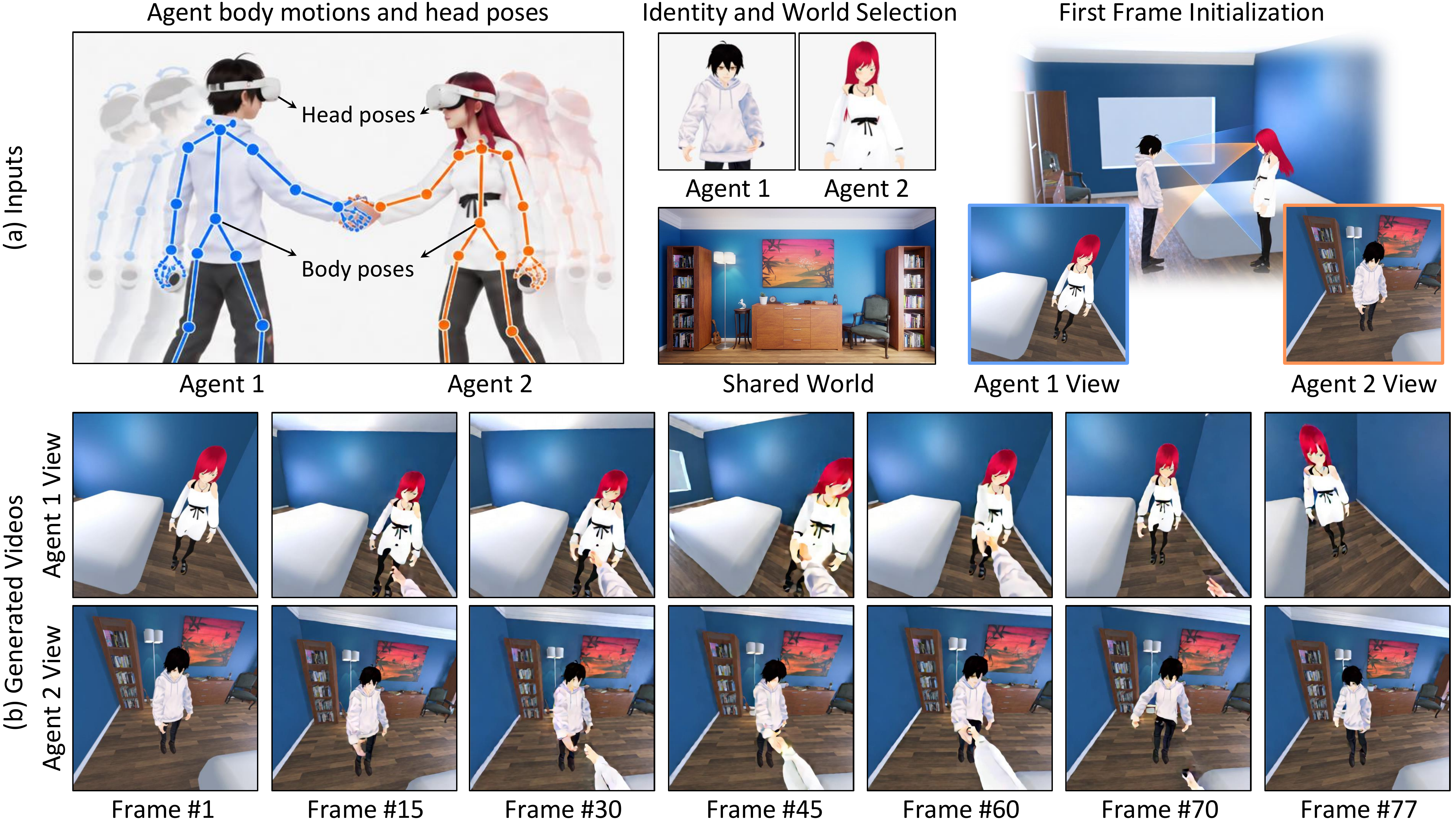}
    \vspace{-20pt}
    \caption{\textbf{Teaser.} Given multi-agent motions, head poses, character identities, and a shared world with first-frame initialization in (a), \textbf{ME-World} jointly generates synchronized ego streams (b) that remain consistent in character identity, scene appearance, and interaction outcomes across agents.}
\label{fig:teaser}
\end{figure}

\begin{abstract}
Egocentric world models predict first-person observations conditioned on an agent’s actions, but most focus on a single agent. Real embodied settings often involve multiple agents that act and interact within a shared environment. Existing multi-agent world models rely on coarse actions like locomotion, camera control, or discrete commands, leaving fine-grained embodied interactions underexplored. We formulate multi-agent egocentric world modeling as synchronized ego-stream generation for multiple agents interacting through fine-grained actions in a shared world. This requires cross-view action consistency, shared-environment consistency, and consistent propagation of interaction-induced state updates. We propose \textbf{Multi-agent Egocentric World Model (ME-World)}, which jointly denoises multiple ego streams in a shared token sequence, conditions each stream on all agents’ target-view poses, and grounds generation with shared environment memory. We train and evaluate on real and synthetic multi-agent data and introduce shared-world consistency metrics for environment, update, and identity consistency. Experiments show \ours improves shared-world consistency, action control, identity preservation, and video quality over existing methods.

\end{abstract}    
\section{Introduction}
\label{sec:intro}
\begin{figure}[t!]       
    \centering    
    \includegraphics[width=\linewidth]{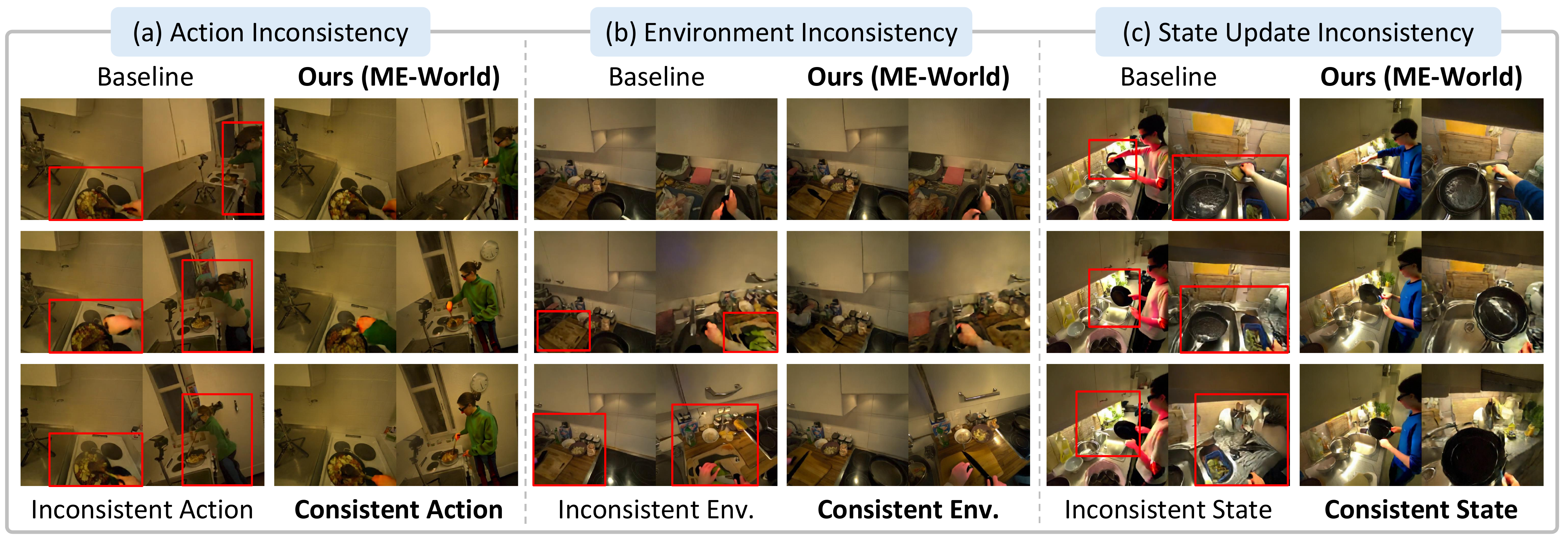}\
    \vspace{-20pt}
    \caption{\textbf{Motivation.} Three challenges in multi-agent egocentric world modeling. Without shared action conditioning, (a) actions are inconsistent across ego streams. Without joint generation, independent ego streams show (b) environment inconsistency and (c) inconsistent state updates.
}
    \vspace{-10pt}
    \label{fig:motivation}   
\end{figure}

Video world models~\citep{ha2018recurrent} generate future observations conditioned on a user's actions or controls, enabling an agent to imagine how a scene may evolve without acting in the real world. Such action-conditioned prediction has been widely explored in autonomous driving~\citep{gao2024vista}, robotics~\citep{kim2026cosmospolicy, ye2026dreamzero}, open-domain world simulation~\citep{nvidia2025cosmosworldfoundationmodel, genie3}, and games~\citep{wang2026matrixgame30, tang2026hunyuangamecraft2}. Recent work has extended it to egocentric world modeling~\citep{zhang2026jointcontrolvideo, hao2026egosim}, where an embodied agent predicts its future first-person observations, or \textit{ego stream}, conditioned on its own actions.

Most existing egocentric world models~\citep{bai2026whole, chen2026handsonworld, gu20263, pallotta2026egocontrol, xiu2026egotwin, li2026anchorworld} assume a single embodied ego, referred to as an \textit{agent}, and capture how its actions shape its own future ego stream. In real-world settings, however, multiple agents often coexist and interact with one another and with shared objects, jointly changing a common, evolving state of the environment, objects, and agents, which we term the \textit{shared world state}. A multi-agent world model must therefore make each interaction visible from every agent's viewpoint and reflect the resulting state changes simultaneously across all observations.

Recent multi-agent egocentric world models~\citep{savva2026solaris, wu2026multiworld, hu2026metaworld} extend egocentric world modeling to multiple agents in a shared environment. However, they mostly target domains where actions are abstracted into low-dimensional controls, such as discrete commands in multiplayer games~\citep{enigma2025multiverse}, camera trajectories in driving and videos~\citep{zhu2026shareverse, sun2026prisma}, and robot control in manipulation~\citep{liu2026gamma}. While such abstractions support multi-view generation under given controls, they fail to capture the dual role of fine-grained embodied actions as both first-person controls and cross-view observable events. In human interaction, body and hand motion, and head-induced viewpoint changes must remain consistent across agents, while the same interaction may also update shared objects and the environment.

We address this gap in \textit{embodied multi-agent settings} as shown in Fig.~\ref{fig:teaser}, which arise naturally in multi-agent generative VR~\citep{tu2026playerone}, embodied AI simulation~\citep{puig2023habitat30}, and multi-robot collaboration~\citep{chen2026rhythm}. In these setting, the challenge goes beyond generating different viewpoints of a common world. \textbf{(i)} Cross-view actions must be synchronized, so that an agent's action appears plausible in its own stream and matches how others observe it (Fig.~\ref{fig:motivation} (a)). \textbf{(ii)} The shared environment must remain consistent, so that ego streams do not depict conflicting layouts, appearances, or objects (Fig.~\ref{fig:motivation} (b)). \textbf{(iii)} Interaction-induced state changes must propagate to every ego stream rather than diverging into conflicting object states or outcomes (Fig.~\ref{fig:motivation} (c)).
We therefore ask: \textit{How can a world model generate multiple ego streams consistent with a single evolving shared world, while each agents observe, act, and interact differently?}

To address these challenges, we introduce \textbf{\underline{M}ulti-agent \underline{E}gocentric \underline{World} Model (ME-World)}, which generates multiple ego streams as coupled observations of one shared world. Rather than generating each stream independently, \ours performs \textit{joint multi-agent generation} by denoising all streams in a shared token sequence, resolving interaction outcomes and state changes consistently across streams. To make this joint generation aware of embodied actions, we introduce \textit{shared action conditioning}, which provides the target agent's own motion and other agents' motion observed from its stream. The generated streams are further grounded by \textit{shared environment memory}, which provides view-aligned evidence by warping history observations into each target view, and common anchor references shared across all streams to maintain scene appearance. By coupling generation, action conditioning, and environment evidence across agents, \ours maintains consistency in cross-view actions, shared environment, and interaction-induced state updates.

Learning this model requires synchronized multi-ego data sharing one world state. Existing egocentric datasets~\citep{grauman2022ego4d,grauman2024egoexo4d,sener2022assembly101, rai2021home, sigurdsson2018charades, damen2018scaling, jia2020lemma, ma2024nymeria} primarily capture single agents, or lack coverage of agent number, viewpoint diversity, interactions, and appearance variation. We therefore combine real multi-ego data~\citep{gavryushin2026comind} with synthetic multi-ego data~\citep{xu2024inter, liang2024intergen} covering diverse agent configurations, environments, and interaction scenarios.

Finally, we introduce an evaluation protocol for \textit{shared-world consistency}. Existing evaluations measure video quality~\citep{unterthiner2019fvd, zhang2018lpips}, control alignment~\citep{gu20263}, and cross-view consistency through geometric agreement or VLM-based judgments~\citep{wu2026multiworld, hu2026metaworld, savva2026solaris}, but do not assess whether multiple ego streams consistently represent the same evolving shared world beyond directly provided input. We therefore propose $S_{\mathrm{env}}$, which measures whether newly generated environment regions remain consistent across ego streams, and $S_{\mathrm{update}}$, which measures whether interaction-induced state changes, such as a moved frying pan, are reflected consistently across views. On the synthetic benchmark, we further report $S_{\mathrm{id}}$ for identity preservation with respect to ground-truth reference characters. These metrics complement quality and control metrics for evaluating embodied multi-agent world modeling.

\ours consistently improves shared-world consistency, action control, identity preservation, and video quality on real and synthetic benchmarks. Ablations validate our three core components, showing the importance of modeling ego streams as coupled observations of one evolving world.

Our contributions are summarized as follows:
\begin{itemize}
    \item We introduce \textbf{embodied multi-agent world modeling} for synchronized ego stream generation among multiple agents interacting in a shared world. We propose \textbf{\ours}, coupling streams via joint generation, shared action, and shared environment memory.

    \item We combine real and synthetic multi-agent data covering diverse environments, appearances, and interactions. We introduce \textbf{shared-world consistency metrics} evaluating whether ego streams depict the same environment and reflect induced state changes.
\end{itemize}

\section{Related Works}
\label{sec:rel_work}

\paragrapht{Egocentric World Models.}
Some egocentric world models condition on whole-body motion~\citep{bai2026whole,tu2026playerone,pallotta2026egocontrol}, fine-grained hand and camera control~\citep{xie2026generatedreality,chen2026handsonworld}, or jointly generated body motion~\citep{xiu2026egotwin}. Others model human-world interaction through dexterous manipulation~\citep{kim2026dexterous,goswami2026dexwm}, explicit 3D memory~\citep{gu20263,yan2026egogenesis}, or goal-directed world evolution~\citep{hao2026egosim,li2026anchorworld,shen2026egoforge}. These focus on a single agent, whereas we model multiple embodied agents whose actions and interaction-induced changes remain consistent across ego views within a shared world.

\paragrapht{Multi-agent World Models.}
Recent video world models extend to multiple agents whose actions and observations are coupled through a shared environment, including multiplayer games~\citep{enigma2025multiverse,savva2026solaris,wu2026multiworld,liu2026gamma,po2026multigen,hu2026multiplayer}, driving~\citep{zhu2026shareverse}, robotics~\citep{wu2026multiworld,liu2026gamma}, and open-domain videos~\citep{hu2026metaworld}. A challenge is maintaining a coherent world across controlled views, addressed via joint or cross-agent attention~\citep{savva2026solaris,sun2026prisma,hu2026metaworld,liu2026gamma,zhu2026shareverse}, shared scene representations~\citep{wu2026multiworld}, or shared memories and explicit world states~\citep{po2026multigen,mo2026streaming,cai2026mass,zhao2026population,xu2026consistworld}. In contrast, we study synchronized egocentric views of interacting humans, where fine-grained articulated actions must remain consistent across ego streams while affecting other agents and the shared environment.

\section{Method}
\label{sec:method}
\begin{figure}[t!]       
    \centering    
    \includegraphics[width=\linewidth]{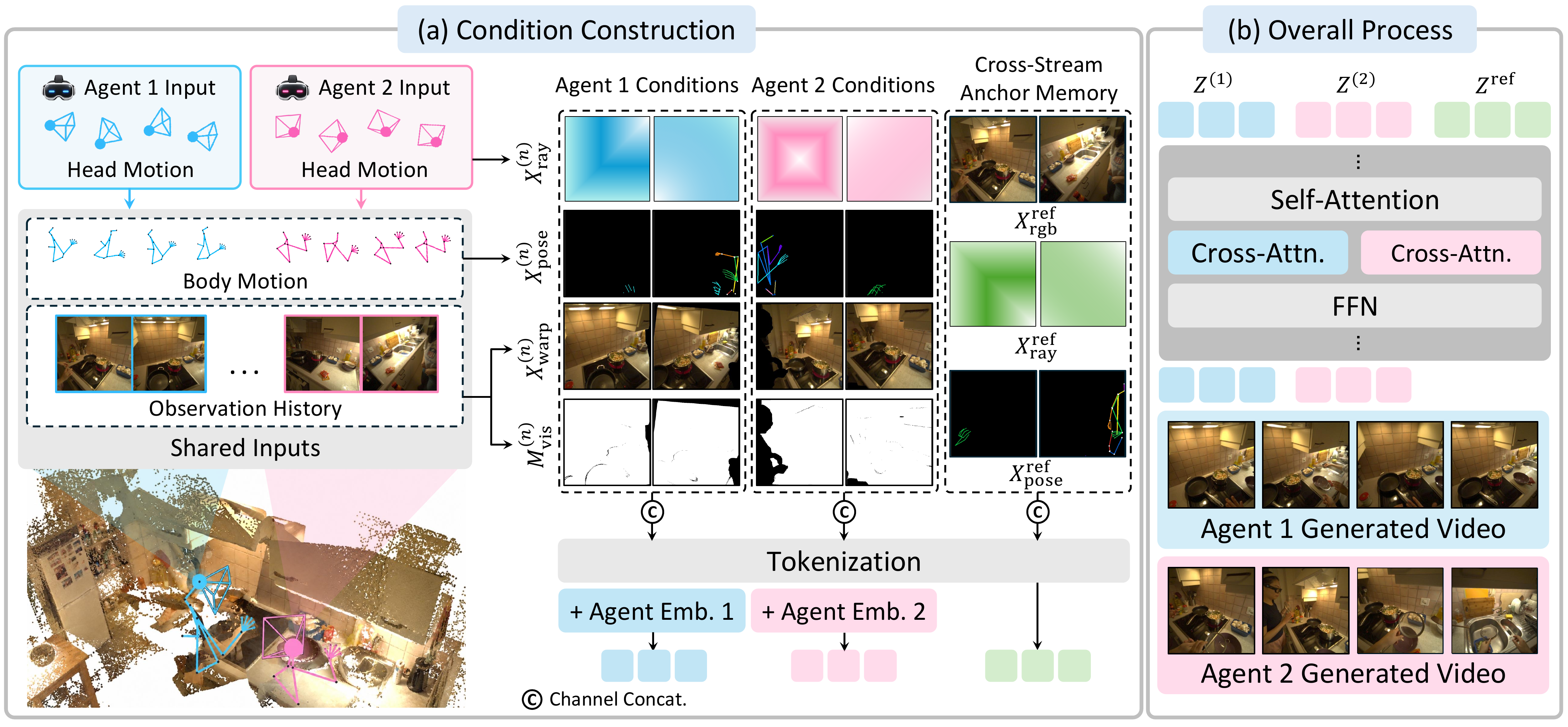}
    \vspace{-20pt}
\caption{
\textbf{Overall architecture}
(a) For each agent, \ours builds viewing ray conditions and shared-action conditions from agents' input and build shared environment memory from the both agents' history.
(b) \ours jointly generates synchronized ego streams.
}
    \vspace{-15pt}
    \label{fig:main_architecture}   
\end{figure}

\subsection{Overview and Formulation}
\textbf{\ours} addresses embodied multi-agent world modeling, where $N$ agents act and interact in the same evolving shared world while observing it from their own egocentric viewpoints. Given each agent's future embodied action sequence, text prompt, initial ego frame and observation history, our goal is to generate future ego streams for all agents, with each stream following its own action and all streams remaining consistent with the same shared environment and shared world state.

For each $n$-th agent, we denote its future ego stream and action sequence which includes head and body motion as $X^{(n)}=\{x_t^{(n)}\}_{t=1}^T$, and $A^{(n)} = \{a_t^{(n)}\}_{t=1}^T$, where $T$ is the length of frames. We also denote its text prompt and initial ego frame as $P^{(n)}$ and $x_0^{(n)}$, respectively. The observation histories of all agents are pooled into shared observation-history pool $E$. The full conditioning signal is defined as $C=\{E, A^{(1:N)}, P^{(1:N)},x_0^{(1:N)} \}$, where $A^{(1:N)}, P^{(1:N)}$, and $x_0^{(1:N)}$ denote the action sequences, text prompts and initial ego frames of all agents, respectively.

The formulation is general to $N$ agents, but for clarity we present the two-agent case used in our experiments. 
The key challenge is that each agent observes only a partial view of the world, while the generated streams must agree on the shared environment, the agents’ actions, and interaction-induced state changes. To address this, \ours couples ego streams through joint multi-agent generation (Sec.~\ref{sec:joint_generation}), guides them with shared action conditioning (Sec.~\ref{sec:shared_action}), and grounds them with shared environment memory (Sec.~\ref{sec:shared_environment}). The overall architecture is shown in Fig.~\ref{fig:main_architecture}.

\subsection{Joint Multi-Agent Generation}
\label{sec:joint_generation}

Since agents interact within the same environment, their ego streams are not independent. When an action changes the state of an agent, object, or region, this update should be reflected in all relevant ego streams rather than resolved separately per stream. We therefore generate all streams jointly within a single pretrained video diffusion transformer (DiT)~\citep{ali2025world}, as shown in Fig.~\ref{fig:main_architecture} (b). Rather than generating each stream independently, the model jointly denoises their latent sequences under the conditioning $C$, allowing cross-stream information exchange. Formally, it learns the joint conditional distribution $p_\theta(X^{(1)}, \dots, X^{(N)} \mid C)$, where $\theta$ denotes the model parameters.

Using a frozen video VAE, each ego stream $X^{(n)}$ is encoded into a latent sequence $Z^{(n)}$. During denoising, all agents' noisy latents at timestep $\tau$ are concatenated along the token dimension as:
\begin{equation}
Z_\tau^{\mathrm{gen}}
=
\left[
Z_\tau^{(1)};\ldots;Z_\tau^{(N)}
\right].
\label{eq:joint_sequence}
\end{equation}

The DiT processes this joint sequence with self-attention, allowing evidence from one ego stream to inform others, so all streams evolve as mutually informed observations of the same shared world rather than isolated videos. Within this joint sequence, however, the model must still identify which stream each token belongs to and which conditions it should follow. We apply identical rotary positional embeddings (RoPE)~\citep{su2023roformer} to all streams, aligning their space-time positions in a common positional frame, and distinguish the streams by adding a learnable agent embedding $e^{(n)}$ to the tokens of $n$-th agent. For text conditioning, each stream uses a separate cross-attention path, since different agents may observe different regions of the same scene. Each stream thus follows its own prompt $P^{(n)}$, while visual information is shared through joint self-attention.

\subsection{Shared Action Conditioning}
\label{sec:shared_action}

Joint generation allows ego streams to exchange information through self-attention, but does not specify \textit{how each agent should move in the shared world}. A straightforward method is to condition each ego stream on its own action, treating the action as private to the stream. However, in multi-agent generation, an agent's action is also observed from other agents' viewpoints and can change the shared world state via interaction. If each stream receives only its own action, the same embodied interaction may be rendered inconsistently across viewpoints. We therefore use \textbf{shared action conditioning}, guiding each ego stream with all agent's motion rather than by its own action alone. 

For each $n$-th agent, the action at frame $t$ is decomposed into body and head motion as $a_t^{(n)}=\{a_{t,\text{body}}^{(n)}, a_{t,\text{head}}^{(n)}\}$. The body motion $a_{t,\text{body}}^{(n)}$ is encoded as a self-action skeleton map capturing ego-visible hand and arm motion. The head motion $a_{t,\text{head}}^{(n)}$ is treated as an action in egocentric generation, since it determines the agent’s camera trajectory and visible region. Head motion is represented with Pl\"ucker ray maps, which provide a geometric encoding of the induced camera rays.

To construct the shared action condition for $n$-th agent's ego stream, we define a pose condition sequence $X_\text{pose}^{(n)} = \{x_{t,\text{pose}}^{(n)}\}_{t=1}^T$. At each frame, $x_{t,\text{pose}}^{(n)}$ is rendered on the target image plane using $n$-th agent's self-action skeleton map $a_{t,\text{body}}^{(n)}$, other agents' 3D body poses, and relative camera geometry. The skeleton map captures the ego-visible own hand and arm motion, while other agents’ poses are projected to $n$-th agent's camera coordinates. This yields a unified pose condition representing both $n$-th agent's own motion and the visible motions of other agents in $n$-th agent's ego view.

For head motion, we collect Pl\"ucker ray maps as $X_\text{ray}^{(n)}=\{a_{t,\text{head}}^{(n)}\}_{t=1}^T$, encoding per-pixel camera rays induced by $n$-th agent's head motion. We express all ray maps in a shared canonical frame, making camera trajectories and viewing directions across ego streams geometrically comparable.

$X_\text{pose}^{(n)}$ is encoded with a frozen VAE encoder, while $X_\text{ray}^{(n)}$ is projected with an MLP. The encoded pose and ray features are concatenated with the noisy latent $Z_\tau^{(n)}$ along the channel dimension and combined with the remaining conditions for joint denoising.
Unlike conditioning each ego stream only on the target agent's own motion, shared action conditioning renders other agents' body poses into the target ego view, giving the model explicit cross-agent motion cues. This is important for fine-grained embodied interaction, where hands, arms, body pose, and gaze-induced viewpoint changes must remain consistent across ego streams. As a result, the model better represents interactions such as approaching, manipulating, or responding to another agent within the same shared world.

\subsection{Shared Environment Memory}
\label{sec:shared_environment}
Shared action conditioning makes the agents’ actions explicit, but it does not tell the model \textit{what the surrounding world should look like}. Even with joint self-attention, each ego stream still observes only a partial view, so the model may lack a common reference for scene layout, background appearance, and object state. We therefore introduce \textbf{shared environment memory} from the shared observation-history pool $E$, which aggregates the observation histories of all agents. 

A straightforward way to use this memory is to feed the selected history images as reference images. However, using the history images as references conditions each stream only implicitly, where it does not specify what each generated frame should depict in that agent's view. We therefore introduce the \textbf{stream-aligned geometric memory} path, which transforms relevant observations in $E$ into target-view conditions for each ego stream, providing geometry-aligned evidence for the frames being generated. Such evidence, however, constrains only the regions it covers and discards the original appearance. The \textbf{cross-stream anchor memory} path complements it by appending selected observations in their clean, unwarped form as shared anchor tokens, anchoring the global appearance and grounding all streams in the same environment during joint denoising.

\paragrapht{Stream-Aligned Geometric Memory.}
The stream-aligned geometric memory path makes the shared environment memory directly usable for each target ego stream. Since history images in $E$ come from past viewpoints, they may not directly match the ego viewpoint. For each target stream, we select the history image $X_\text{hist}$ whose reprojection gives the largest valid coverage of the target view, and warp it according to $n$-agent's target ray condition $X_\text{ray}^{(n)}$. During projection, we mask dynamic agent regions using SAM3~\citep{carion2026sam} to avoid transferring outdated poses. This warping yields a warped RGB condition $X_\text{warp}^{(n)}$ and a visibility mask $M_\text{vis}^{(n)}$, which provide aligned scene evidence and indicate supported regions, respectively. We encode these conditions with a frozen VAE encoder and concatenate them with the noisy latent $Z_\tau^{(n)}$ along the channel dimension, directly conditioning denoising on the aligned scene evidence. 

\paragrapht{Cross-Stream Anchor Memory.}
While stream-aligned geometric memory provides target-view guidance, it is inherently view-specific so that each stream is conditioned only on evidence that can be reliably projected into its own viewpoint, rather than on a holistic view of the scene. This can leave regions missing, discard clean appearance cues from the original observations, and limit the model's understanding of the shared environment as a whole. To complement this, the cross-stream anchor memory path preserves selected history images from $E$ in their original, unwarped form as shared scene references. We greedily select $K$ history observations, choosing each new anchor to maximize the target-view regions not covered by previously selected anchors, so that the anchors provide complementary scene coverage. The selected observations are encoded by the VAE into reference features $Z_\text{ref}=\{z_\text{ref}^k\}_{k=1}^K$ and appended to the joint denoising sequence along the token dimension, allowing all ego streams to attend to the same anchors during generation. Extending Eq.~\ref{eq:joint_sequence} with shared environment memory, the input joint sequence to the DiT is formed as:

\[
Z_\tau^{\mathrm{joint}}
=
\left[
Z_\tau^\text{gen};
Z_{\mathrm{ref}}
\right],
\]
where each \({Z}_\tau^{(n)}\) is channel-wise concatenated with shared action and stream-aligned conditions, and \(Z_{\mathrm{ref}}\) denotes the cross-stream anchor features augmented with corresponding camera-ray and shared action conditions.

\subsection{Training Objective}
\label{sec:training_objective}
We fine-tune the pretrained model with rectified flow~\citep{liu2022flow}.
Let \(Z_0^{\mathrm{gen}}=[Z_0^{(1)};\ldots;Z_0^{(N)}]\) denote the clean latents of all agents, and \(\epsilon^{\mathrm{gen}}=[\epsilon^{(1)};\ldots;\epsilon^{(N)}]\), where \(\epsilon^{(n)}\sim\mathcal{N}(0,I)\).
For \(\tau \sim \mathcal{U}(0,1)\), we construct \(Z_\tau^{\mathrm{gen}}=(1-\tau)Z_0^{\mathrm{gen}}+\tau\epsilon^{\mathrm{gen}}\).
The model $f_\theta$ is trained with
\[
\mathcal{L}_{\mathrm{FM}}
=
\mathbb{E}_{Z_0,\epsilon,\tau,C}
\left[
\left\|
f_\theta
\left(
Z_\tau^{\mathrm{joint}}, \tau, C
\right)
-
\left(
\epsilon^{\mathrm{gen}} - Z_0^{\mathrm{gen}}
\right)
\right\|_2^2
\right].
\]

\section{Experiments}
\subsection{Shared-World Consistency Metric}
\label{sec:shared_world_metric}
\begin{figure}[t!]       
    \centering    
    \includegraphics[width=\linewidth]{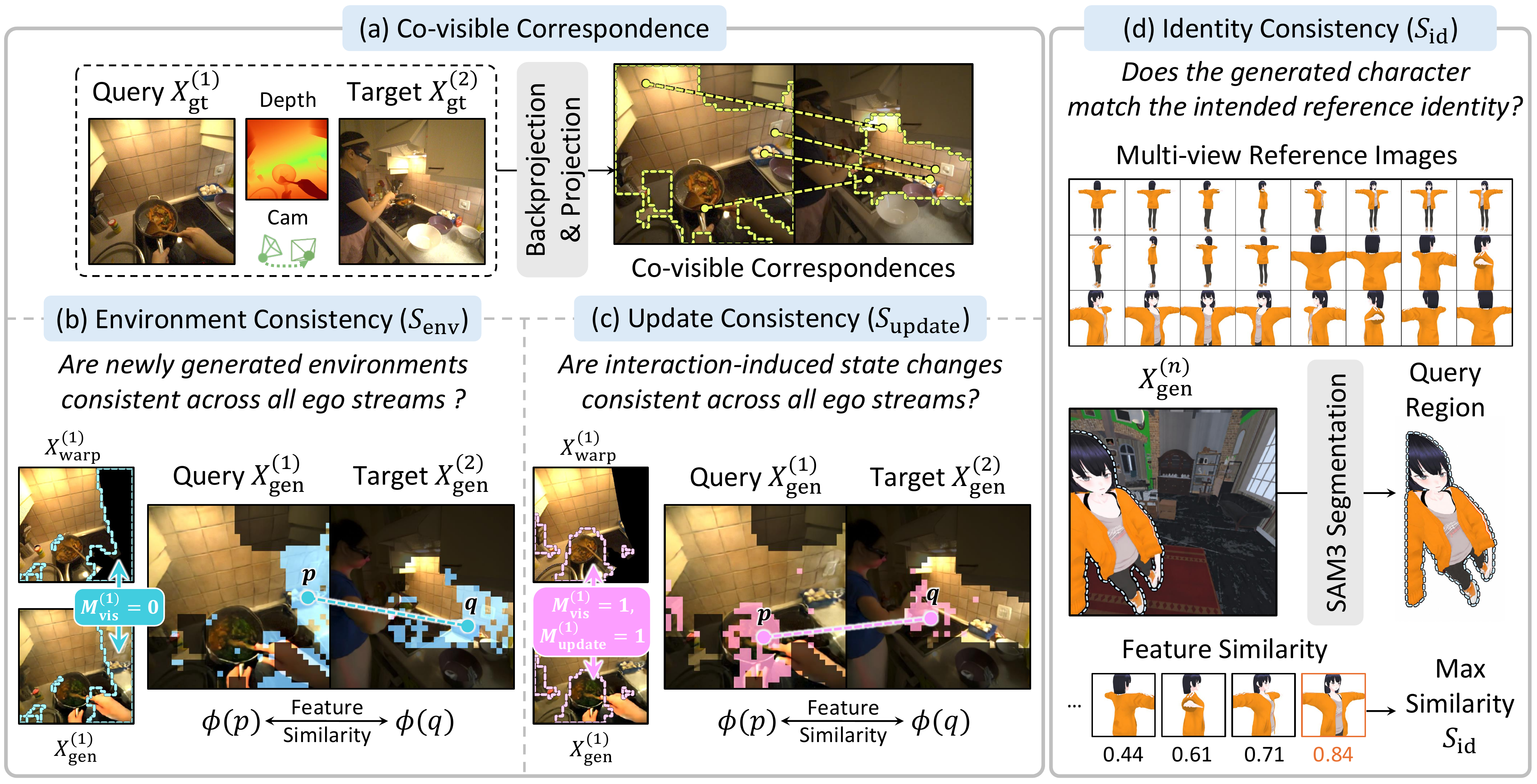}
    \vspace{-20pt}
\caption{
\textbf{Evaluation Protocol for Shared-World Consistency.}
We evaluate generated environments with $S_{\mathrm{env}}$, state updates with $S_{\mathrm{update}}$, and identity with $S_{\mathrm{id}}$.
Ground-truth geometry is used only to establish co-visible correspondences, while consistency is measured on generated streams.
}
    \vspace{-15pt}
    \label{fig:eval_protocol}   
\end{figure}

Existing multi-agent world models evaluate cross-view consistency using geometric agreement or VLM-based judgments~\citep{wu2026multiworld,hu2026metaworld, savva2026solaris}. However, such consistency can arise directly from shared input conditions, making it difficult to isolate the consistency established by the model itself. Distinguishing this consistency is critical for assessing whether the model can maintain a shared environment beyond the explicit input. We therefore introduce $S_\text{env}$, which measures environment consistency only in regions not directly provided by geometric memory. Beyond the environment itself, a coherent shared world should consistently reflect interaction-induced state changes across agent views, which we measure with $S_\text{update}$. It should also preserve the identity of each agent when observed from different viewpoints, which we measure with $S_\text{id}$.

For evaluation, we denote the $n$-th agent's ground-truth and generated stream as $X_{\mathrm{gt}}^{(n)}$ and $X_{\mathrm{gen}}^{(n)}$, respectively. Fig.~\ref{fig:eval_protocol} (a) visualizes, on ground-truth views for clarity, the co-visible correspondences used for $S_{\mathrm{env}}$ and $S_{\mathrm{update}}$, obtained from metric depth and ground-truth camera poses. These scores are computed on generated stream pairs. We report $S_\text{env}$ and $S_\text{update}$ on the real dataset, and all three metrics on synthetic dataset (Appendix~\ref{sec:supp_synthetic_dataset}), where distinct character appearances enable reference-based $S_\text{id}$ evaluation. Additional details are provided in Appendix~\ref{sec:supp_shared_world_metric_main}. For clarity, we describe metrics using 1st agent as the query stream and 2nd agent as the paired stream.

\paragrapht{Environment Consistency.}
$S_\text{env}$ evaluates whether the model maintains a shared environment in regions unsupported by warped history. For a query stream $X_\text{gen}^{(1)}$, we select regions where $M_\text{vis}^{(1)}=0$, meaning $X_\text{warp}^{(1)}$ provides no valid scene guidance. We measure DINOv3 feature similarity between these regions and corresponding regions in $X_\text{gen}^{(2)}$ defined by co-visible correspondence, since they should depict the same shared environment despite lacking warped-history guidance.

\paragrapht{State Update Consistency.}
$S_\text{update}$ evaluates whether regions updated from the warped history are consistent across ego streams. We define an update mask $M_\text{update}^{(1)}$ by comparing $X_\text{gen}^{(1)}$ with $X_\text{warp}^{(1)}$ in feature space, where high feature difference indicates generated content that has changed from the history. For query stream $X_\text{gen}^{(1)}$, we select regions with valid warped-history guidance and detected updates, where $M_\text{vis}^{(1)} =1$ and $M_\text{update}^{(1)}=1$. We measure DINOv3 feature similarity~\citep{simeoni2025dinov3} between these updated regions and the corresponding regions in $X_\text{gen}^{(2)}$ defined by co-visible correspondences, since interaction-induced changes should appear as shared-world updates.

\paragrapht{Identity Consistency.}
On the synthetic benchmark, $S_\text{id}$ evaluates whether the visible other agent preserves its intended appearance as part of the shared world. For each generated stream, we extract the visible other agent using SAM3~\citep{carion2026sam} and compute its DINOv3 feature~\citep{simeoni2025dinov3}. We then compare it with the multi-view reference features of the corresponding ground-truth character and use the maximum cosine similarity as the identity score. This measures whether different ego streams depict the same counterpart in the shared world, rather than drifting to inconsistent character identities. The final score is averaged over valid frames and ego streams.

\begin{figure}[t!]       
    \centering    
    \includegraphics[width=\linewidth]{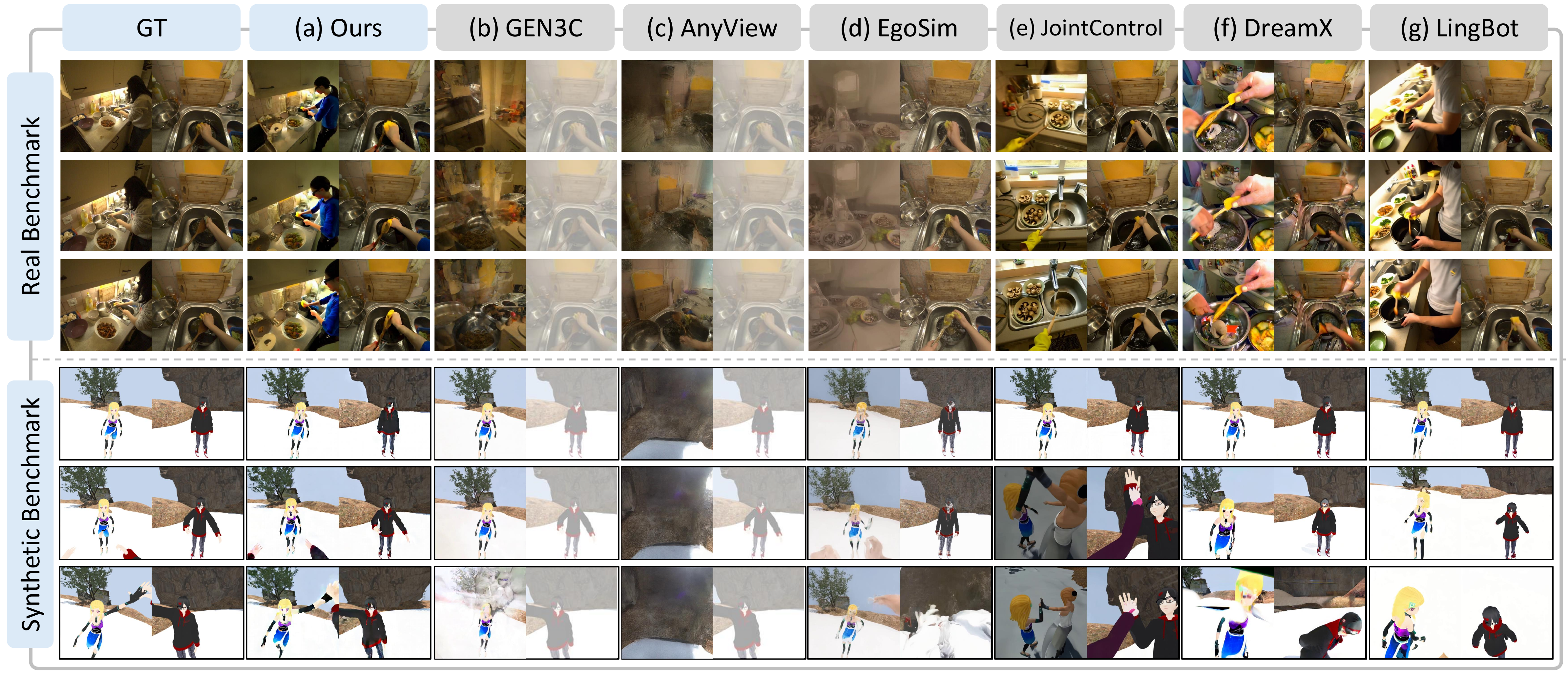}
    \vspace{-20pt}
\caption{
\textbf{Qualitative Comparison with Existing Methods.}
We compare \ours on the real (top) and synthetic (bottom) benchmarks.
Each column shows three timestamps from two synchronized ego streams.
For multi-view baselines, the ground-truth conditioning stream is shown with reduced opacity. Best viewed zoomed in.
}
    \vspace{-11pt}
    \label{fig:main_qual}   
\end{figure}

\begin{figure}[!tb]
    \centering
    \includegraphics[width=\linewidth]{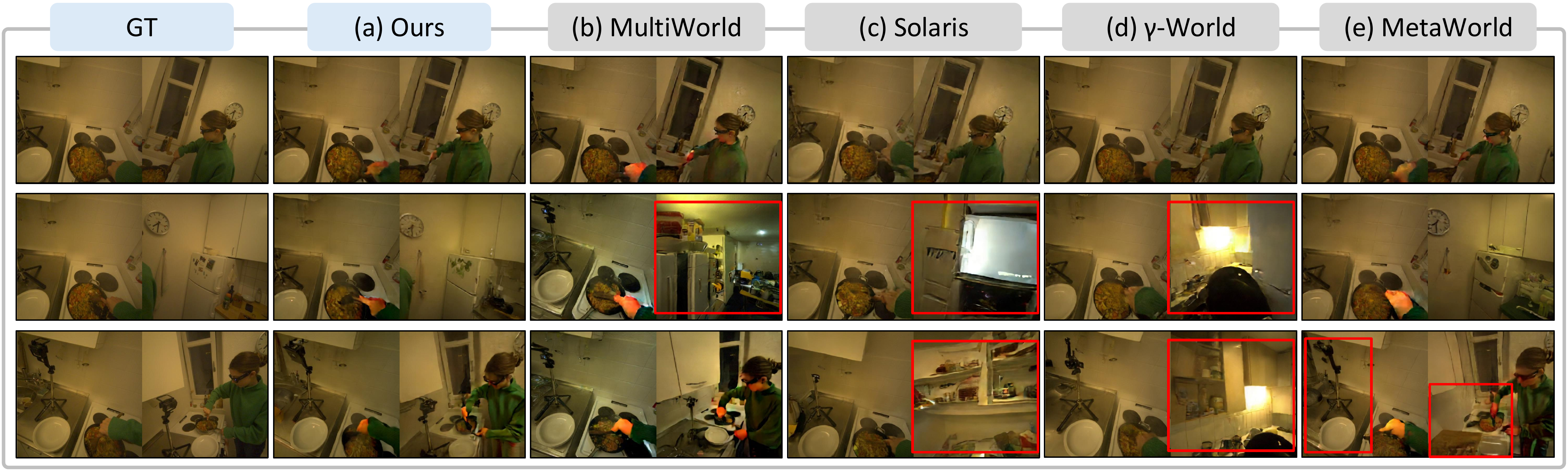}
    \vspace{-20pt}
\caption{
\textbf{Qualitative Comparison with Multi-Agent World Models.}
We compare \ours with existing multi-agent world models on the real benchmark.
Each column shows three timestamps from two synchronized ego streams.
\ours better preserves a coherent shared environment and synchronized interactions across the two viewpoints.
Best viewed zoomed in.
}
    \vspace{-15pt}
    \label{fig:main_qual_multiagent}
\end{figure}
\subsection{Implementation Details}
\paragrapht{Model and Training Setup.}
We initialize from the Cosmos-Predict2.5 2B\citep{ali2025world}
multi-view diffusion transformer. We train separate models on  
fine-tuning the DiT and conditioning modules while keeping the video VAE and text encoder frozen. We use the rectified-flow objective with AdamW~\citep{loshchilov2019adamw}, a learning rate of $3\times10^{-5}$, and a global batch size of 4 on four A100 GPUs.
At inference, we use 35 denoising steps and a classifier-free guidance scale of 7.0. Further details are provided in Appendix~\ref{sec:supp_training_details}.

\paragrapht{Evaluation Metrics.}
We also evaluate camera control, self-action control, other-agent action control, and video quality. For camera control, we use Translation Error (Trans. Err) and Rotation Error (Rot. Err). Following E$^3$C~\citep{gu20263}, self-action control measures the  agent's hand presence and alignment using Hand-F1 and Hand-mIoU, while other-agent action control measures the visibility and pose alignment using Full-Body-F1 and PCK@10. For video quality, we report PSNR~\citep{huynh2008psnr}, SSIM~\citep{wang2004ssim}, LPIPS~\citep{zhang2018lpips}, and FVD~\citep{unterthiner2019fvd}. Additional details are provided in Appendix~\ref{sec:supp_eval_metrics}.

\paragrapht{Baselines.}
We compare three categories of methods:
(i) multi-view video generation models (GEN3C~\citep{ren2025gen3c}, AnyView-DVS~\citep{van2026anyview}),
(ii) single-ego world models (EgoSim~\citep{hao2026egosim}, JointControlVideo~\citep{zhang2026jointcontrolvideo}),
and (iii) general world models (DreamX-World~\citep{team2026dreamx}, LingBot-World~\citep{team2026advancing}).
All baselines use released checkpoints without fine-tuning and receives their supported conditions. Multi-view models use ground-truth source videos, whereas single-ego and general world models generate the two streams independently. Implementations details are provided in the Appendix~\ref{sec:supp_baseline_details}. 

\paragrapht{Multi-Agent Baselines.}
To further compare with recent multi-agent world models under a controlled setting, we adapt MultiWorld~\citep{wu2026multiworld}, Solaris~\citep{savva2026solaris}, $\gamma$-World~\citep{liu2026gamma}, and MetaWorld~\citep{hu2026metaworld} to our embodied multi-agent setting. All adaptations use the same backbone, training data, and Shared Action Conditioning as ME-World, while retaining the characteristic multi-agent and scene-conditioning mechanisms of each method. Implementation details provided in Appendix~\ref{sec:supp_adaptation_details}.

\subsection{Results}
\begin{table*}[t]
\centering
\caption{
\textbf{Quantitative Comparison with Existing Methods.}
We compare \ours with multi-view generation,
single-ego world, and general world models
on the real and synthetic benchmarks.
N/A denotes metrics that are not applicable to a benchmark.
}
\resizebox{\textwidth}{!}{
\begin{tabular}{l|ccc|cc|cc|cc|cccc}
\toprule
&
\multicolumn{3}{c|}{\textbf{Shared-World Consistency}}
& \multicolumn{2}{c|}{\textbf{Camera Control}}
& \multicolumn{2}{c|}{\textbf{Self Action Control}}
& \multicolumn{2}{c|}{\textbf{Other-Agent Action Control}}
& \multicolumn{4}{c}{\textbf{Video Quality}} \\
\cmidrule(lr){2-4}
\cmidrule(lr){5-6}
\cmidrule(lr){7-8}
\cmidrule(lr){9-10}
\cmidrule(lr){11-14}
Method
& $S_{\mathrm{env}}\uparrow$
& $S_{\mathrm{update}}\uparrow$
& $S_{\mathrm{id}}\uparrow$
& Trans. Err.$\downarrow$
& Rot. Err.$\downarrow$
& Hand-F1$\uparrow$
& Hand-mIoU$\uparrow$
& Full-Body-F1$\uparrow$
& Full-Body-PCK$\uparrow$
& PSNR$\uparrow$
& SSIM$\uparrow$
& LPIPS$\downarrow$
& FVD$\downarrow$ \\

\midrule
\rowcolor{gray!15}
\multicolumn{14}{c}{\textit{Real Benchmark}} \\
\midrule

GEN3C~\citep{ren2025gen3c}
& \second{0.417} & \second{0.412} & N/A
& 0.12 & 12.376
& 0.537 & 0.087 & 0.229 & 0.456
& \second{18.083} & \second{0.582} & \second{0.444} & 1358 \\

AnyView-DVS~\citep{van2026anyview}
& 0.332 & 0.319 & N/A
& \second{0.079} & \second{4.736}
& 0.169 & 0.005 & 0.037 & 0.071
& 15.899 & 0.546 & 0.603 & 1834.9 \\

\midrule

EgoSim~\citep{hao2026egosim}
& 0.332 & 0.319 & N/A
& 0.114 & 11.501
& 0.75 & \second{0.393} & 0.274 & \second{0.514}
& 16.398 & 0.574 & 0.448 & 1012.7 \\

JointControlVideo~\citep{zhang2026jointcontrolvideo}
& 0.328 & 0.291 & N/A
& 0.166 & 21.663
& \second{0.799} & 0.249 & 0.215 & 0.402
& 15.18 & 0.473 & 0.574 & 694 \\

\midrule

DreamX-World~\citep{team2026dreamx}
& 0.306 & 0.279 & N/A
& 0.145 & 15.979
& 0.754 & 0.06 & 0.342 & 0.274
& 14.045 & 0.442 & 0.618 & 1098.9 \\

LingBot-World~\citep{team2026advancing}
& 0.329 & 0.284 & N/A
& 0.105 & 13.173
& 0.738 & 0.064 & \second{0.395} & 0.243
& 14.8421 & 0.489 & 0.576 & \second{689.9} \\

\midrule

\textbf{\ours (Ours)}
& \best{0.468}
& \best{0.466}
& N/A
& \best{0.042}
& \best{2.256}
& \best{0.888}
& \best{0.531}
& \best{0.822}
& \best{0.881}
& \best{19.849}
& \best{0.656}
& \best{0.289}
& \best{637.9} \\

\midrule
\rowcolor{gray!15}
\multicolumn{14}{c}{\textit{Synthetic Benchmark}} \\
\midrule

GEN3C~\citep{ren2025gen3c}
& \second{0.414} & \second{0.406} & 0.433
& 0.438 & 21.459
& 0.286 & 0.051 & 0.307 & 0.226
& \second{15.032} & \second{0.572} & \second{0.488} & 1299.3 \\

AnyView-DVS~\citep{van2026anyview}
& 0.402 & 0.383 & 0.238
& \best{0.23} & \best{9.004}
& 0.014 & 0.0 & 0.064 & 0.038
& 13.446 & 0.498 & 0.667 & 1523 \\

\midrule

EgoSim~\citep{hao2026egosim}
& 0.318 & 0.301 & 0.43
& 0.363 & 18.193
& 0.441 & 0.095 & \second{0.387} & \second{0.278}
& 14.235 & 0.526 & 0.534 & 1205.7 \\

JointControlVideo~\citep{zhang2026jointcontrolvideo}
& 0.306 & 0.27 & \second{0.518}
& 0.513 & 29.977
& \second{0.527} & \second{0.12} & 0.315 & 0.131
& 11.706 & 0.462 & 0.613 & 886.1 \\

\midrule

DreamX-World~\citep{team2026dreamx}
& 0.38 & 0.358 & 0.484
& 0.413 & 22.644
& 0.26 & 0.016 & 0.347 & 0.127
& 12.17 & 0.456 & 0.599 & 1169.5 \\

LingBot-World~\citep{team2026advancing}
& 0.249 & 0.21 & 0.5
& 0.463 & 29.962
& 0.271 & 0.01 & 0.316 & 0.09
& 10.577 & 0.489 & 0.65 & \second{841.5} \\

\midrule

\textbf{\ours (Ours)}
& \best{0.441} & \best{0.436} & \best{0.561}
& \second{0.293} & \second{9.057}
& \best{0.593} & \best{0.587}
& \best{0.506} & \best{0.644}
& \best{16.595} & \best{0.621}
& \best{0.328} & \best{465.5} \\

\bottomrule
\end{tabular}
}
\vspace{-15pt}
\label{tab:quan}
\end{table*}
\begin{table*}[!tb]
\centering
\caption{
\textbf{Quantitative Comparison with Multi-Agent World Models.}
We compare \ours with multi-agent world models on the real benchmark. Since existing multi-agent world models are designed for different domains and control interfaces,
we adapt them to our embodied multi-agent setting for a controlled comparison.
}
\label{tab:adapted_quan}

\resizebox{\textwidth}{!}{
\begin{tabular}{l|cc|cc|cc|cc|cccc}
\toprule
&
\multicolumn{2}{c|}{\textbf{Shared-World Consistency}}
& \multicolumn{2}{c|}{\textbf{Camera Control}}
& \multicolumn{2}{c|}{\textbf{Self Action Control}}
& \multicolumn{2}{c|}{\textbf{Other-Agent Action Control}}
& \multicolumn{4}{c}{\textbf{Video Quality}} \\
\cmidrule(lr){2-3}
\cmidrule(lr){4-5}
\cmidrule(lr){6-7}
\cmidrule(lr){8-9}
\cmidrule(lr){10-13}

\textbf{Method}
& $S_{\mathrm{env}}\uparrow$
& $S_{\mathrm{update}}\uparrow$
& Trans. Err.$\downarrow$
& Rot. Err.$\downarrow$
& Hand-F1$\uparrow$
& Hand-mIoU$\uparrow$
& Full-Body-F1$\uparrow$
& Full-Body-PCK$\uparrow$
& PSNR$\uparrow$
& SSIM$\uparrow$
& LPIPS$\downarrow$
& FVD$\downarrow$ \\
\midrule

\rowcolor{gray!15}
\multicolumn{13}{c}{\textit{Real Benchmark}} \\
\midrule

MultiWorld~\citep{wu2026multiworld}
& 0.395
& 0.376
& 0.113
& 11.616
& \second{0.868}
& \second{0.497}
& 0.804
& \second{0.837}
& 15.323
& 0.481
& 0.525
& \second{694} \\

Solaris~\citep{savva2026solaris}
& 0.324
& 0.293
& 0.171
& 27.18
& 0.709
& 0.048
& 0.146
& 0.236
& 14.98
& 0.474
& 0.594
& 984.2 \\

$\gamma$-World~\citep{liu2026gamma}
& 0.304
& 0.266
& 0.181
& 24.414
& 0.758
& 0.064
& 0.188
& 0.214
& 14.892
& 0.469
& 0.584
& 788 \\

MetaWorld~\citep{hu2026metaworld}
& \second{0.421}
& \second{0.413}
& \second{0.054}
& \second{4.086}
& 0.865
& 0.379
& \second{0.816}
& 0.787
& \second{18.928}
& \second{0.627}
& \second{0.329}
& 704.4 \\

\midrule

\textbf{\ours (Ours)}
& \best{0.468}
& \best{0.466}
& \best{0.042}
& \best{2.256}
& \best{0.888}
& \best{0.531}
& \best{0.822}
& \best{0.881}
& \best{19.849}
& \best{0.656}
& \best{0.289}
& \best{637.9} \\

\bottomrule
\end{tabular}
}
\vspace{-12pt}
\end{table*}
\paragrapht{Real Benchmark.}
Tab.~\ref{tab:quan} and Fig.~\ref{fig:main_qual} compare \ours with multi-view generation models~\citep{ren2025gen3c, van2026anyview}, single-ego world models~\citep{hao2026egosim, zhang2026jointcontrolvideo}, and general world models~\citep{team2026dreamx, team2026advancing} on the real~\citep{gavryushin2026comind} and synthetic benchmarks. Multi-view methods use one agent’s ground-truth ego stream as a reference, but still show lower shared-world consistency and blurry or ghosted actions, indicating that reference-stream synthesis alone cannot resolve interactions across agents. Single-ego world models follow the target agent’s action, but lack conditioning for the other agent, leading to weaker other-agent action control and mismatched motions under dynamic viewpoints. General world models generate plausible individual streams, but often fail to keep the two ego streams consistent in environment and interaction outcomes. In contrast, \ours consistently improves shared-world consistency, self- and other-agent action control, and video quality while producing coherent scenes and synchronized interactions. Additional results are in Appendix~\ref{sec:supp_additional_qualitative}.

\paragrapht{Synthetic Benchmark.}
The synthetic benchmark enables reference-based evaluation of counterpart identity preservation with $S_{\mathrm{id}}$, as each agent is assigned a distinct character appearance. Existing methods often fail to preserve the appearance and produce mismatched motions across ego streams, causing identity drift and inconsistent interactions. In contrast, \ours achieves the strongest shared-world consistency, improves camera control, self- and other-agent action control, and video quality. Additional synthetic results, including broader multi-agent and autoregressive generation settings, are provided in Appendix~\ref{sec:supp_additional_qualitative} and~\ref{sec:supp_appqual}.

\begin{wraptable}{r}{0.6\linewidth}
\vspace{-20pt}
\centering
\caption{\textbf{Adapted Multi-Agent Architectures.}
$\checkmark$ uses the corresponding ME-World component, $\triangle$ denotes a method-specific replacement, and $\times$ no explicit counterpart.}
\label{tab:adapted_architecture}

\small
\setlength{\tabcolsep}{4pt}

\resizebox{\linewidth}{!}{
\begin{tabular}{l|ccc}
\toprule
\rowcolor{gray!15}
\textbf{Method}
& \shortstack{\textbf{Joint Multi-Agent}\\\textbf{Generation}}
& \shortstack{\textbf{Shared Action}\\\textbf{Conditioning}}
& \shortstack{\textbf{Shared Environment}\\\textbf{Memory}} \\
\midrule

MultiWorld~\citep{wu2026multiworld}
& $\checkmark$
& $\checkmark$
& $\triangle$ \\

Solaris~\citep{savva2026solaris}
& $\checkmark$
& $\checkmark$
& $\times$ \\

$\gamma$-World~\citep{liu2026gamma}
& $\triangle$
& $\checkmark$
& $\times$ \\

MetaWorld~\citep{hu2026metaworld}
& $\triangle$
& $\checkmark$
& $\triangle$ \\

\midrule
\textbf{\ours}
& $\checkmark$
& $\checkmark$
& $\checkmark$ \\

\bottomrule
\end{tabular}
}

\vspace{-10pt}
\end{wraptable}

\paragrapht{Multi-Agent Baselines.}
Tab.~\ref{tab:adapted_quan} and Fig.~\ref{fig:main_qual_multiagent} compare \ours with multi-agent world models. For a controlled comparison in our embodied egocentric setting, we adapt each method using the same video backbone~\cite{nvidia2025cosmosworldfoundationmodel}, training data~\cite{gavryushin2026comind}, and Shared Action Conditioning as \ours.
MultiWorld~\citep{wu2026multiworld} and MetaWorld~\citep{hu2026metaworld} achieve relatively strong action control, with MetaWorld also showing the strongest shared-world consistency and video quality among the baselines. In contrast, Solaris~\citep{savva2026solaris} and $\gamma$-World~\citep{liu2026gamma} show larger degradation in shared-world consistency and fine-grained action control under our embodied setting. \ours achieves the best performance across all metrics, demonstrating stronger shared-world consistency, action control, and video quality.
Tab.~\ref{tab:adapted_architecture} summarizes the resulting adaptations. MultiWorld replaces our Shared Environment Memory with its VGGT-based Global State Encoder, while Solaris uses no explicit environment-conditioning module. $\gamma$-World replaces joint full self-attention with Sparse Hub Attention and Simplex Rotary Agent Encoding, whereas MetaWorld uses World-State Alignment with warped observations and depth-based conditions. We reimplement MetaWorld World-State Alignment following the paper, as no official implementation is publicly available.

\subsection{Ablation Study}
\begin{table*}[t]
\centering
\caption{
\textbf{Component Ablation Results.}
We evaluate the effects of each component. 'FT' means fine-tuned by our real dataset.
}
\resizebox{\textwidth}{!}{
\begin{tabular}{c|l|c|cc|cc|cc|cc|cccc}
\toprule
& &
& \multicolumn{2}{c|}{\textbf{Shared-World Consistency}}
& \multicolumn{2}{c|}{\textbf{Camera Control}}
& \multicolumn{2}{c|}{\textbf{Self Action Control}}
& \multicolumn{2}{c|}{\textbf{Other-Agent Action Control}}
& \multicolumn{4}{c}{\textbf{Video Quality}} \\
\cmidrule(lr){4-5}
\cmidrule(lr){6-7}
\cmidrule(lr){8-9}
\cmidrule(lr){10-11}
\cmidrule(lr){12-15}
Index
& Variant
& FT
& $S_{\mathrm{env}}\uparrow$
& $S_{\mathrm{update}}\uparrow$
& Trans. Err.$\downarrow$
& Rot. Err.$\downarrow$
& Hand-F1$\uparrow$
& Hand-mIoU$\uparrow$
& Full-Body-F1$\uparrow$
& Full-Body-PCK$\uparrow$
& PSNR$\uparrow$
& SSIM$\uparrow$
& LPIPS$\downarrow$ 
& FVD$\downarrow$\\
\midrule

\rowcolor{gray!15}
\multicolumn{15}{c}{\textit{Real Benchmark}} \\
\midrule

(1) & Cosmos-Predict2.5 (I2V + Text)
& $\times$
& 0.289 & 0.245
& 0.178 & 19.993
& 0.783 & 0.080
& 0.248 & 0.284
& 14.701 & 0.468 & 0.586 & 1006 \\

(2) & Cosmos-Predict2.5 (I2V + Text)
& $\checkmark$
& 0.290 & 0.249
& 0.170 & 24.513
& 0.789 & 0.079
& 0.238 & 0.309
& 14.941 & 0.473 & 0.602 & 768.3  \\

\midrule

(3) & (2) + Self Hand Pose + History Warp
& $\checkmark$
& 0.401 & 0.388
& 0.056 & 4.998
& 0.857 & 0.457
& 0.580 & 0.385
& 18.835 & 0.628 & 0.349 & 558.6\\

(4) & (3) + Joint Gen.
& $\checkmark$
& 0.426 & 0.421
& 0.067 & 6.451
& 0.846 & 0.354
& 0.563 & 0.416
& 18.985 & 0.637 & 0.344 & \second{554.7} \\

(5) & (4) + Shared Action
& $\checkmark$
& 0.445 & 0.442
& 0.058 & 4.720
& 0.876 & 0.410
& 0.803 & 0.809
& 19.315 & 0.649 & 0.328 & 651.9\\

(6) & (4) + Shared Env.
& $\checkmark$
& \second{0.456} & \second{0.455}
& \second{0.042} & 2.930
& 0.859 & \second{0.491}
& 0.685 & 0.428
& \second{19.366} & 0.646 & \second{0.299} & \best{541.7}\\

(7) & (3) + Shared Action + Shared Env.
& $\checkmark$
& 0.433 & 0.422
& \best{0.039} & \second{2.565}
& \second{0.879} & 0.489
& \second{0.813} & \second{0.847}
& 19.180 & \second{0.651} & 0.324 & 717.8\\

\midrule

\textbf{(8)} & \textbf{(7) + Joint Gen. (Ours)}
& $\checkmark$
& \best{0.468}
& \best{0.466}
& \second{0.042}
& \best{2.256}
& \best{0.888}
& \best{0.531}
& \best{0.822}
& \best{0.881}
& \best{19.849}
& \best{0.656}
& \best{0.289}
& 637.9\\

\bottomrule
\end{tabular}
}
\vspace{-10pt}
\label{tab:component_abl}
\end{table*}
\begin{figure}[t!]       
    \centering    
    \includegraphics[width=\linewidth]{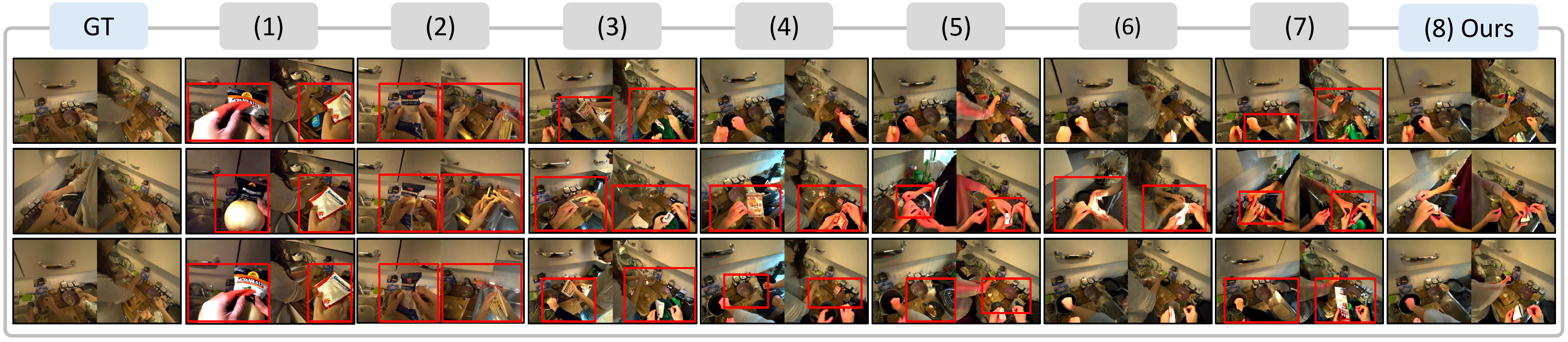}
    \vspace{-20pt}
\caption{
\textbf{Qualitative Results of Ablation Studies.}
Removing individual components leads to inconsistencies in agent appearance, embodied actions, and the shared environment across ego streams.
Red boxes highlight representative artifacts.
Best viewed zoomed in.
}
    \vspace{-15pt}
    \label{fig:abl_qual}   
\end{figure}

Tab.~\ref{tab:component_abl} and Fig.~\ref{fig:abl_qual} show that each component addresses a distinct failure mode. Off-the-shelf and directly fine-tuned Cosmos-Predict2.5 baselines, variants (1) and (2), fail to produce plausible multi-agent ego futures, highlighting the need for explicit multi-agent conditioning.
Without \textbf{joint multi-agent generation}, variants (3) and (7) produce inconsistent ego streams, where agent appearance and the surrounding environment diverge across views.
Without \textbf{shared action conditioning}, variants (3), (4), and (6) exhibit cross-view action misalignment.
Without \textbf{shared environment memory}, variants (3), (4), and (5) generate inconsistent backgrounds across streams.
These results show that joint generation couples the predictions, shared action conditioning aligns embodied behavior, and shared environment memory anchors both streams to a common evolving world.

\section{Conclusion}
\label{sec:conclusion}
We introduced multi-agent egocentric world modeling, where multiple agents interact in a shared world through their own egocentric viewpoints. We proposed \textbf{\ours}, which jointly generates ego streams with shared action conditioning and environment memory. We also constructed real and synthetic benchmarks with shared-world consistency metrics. Experiments show that \ours improves shared-world consistency and fine-grained action control while maintaining strong video quality, highlighting the importance of modeling ego streams as coupled observations of an evolving world.
Our framework offers several directions for extension. Jointly modeling actions and observations could enable agents to plan behaviors and anticipate their effects on the shared world, while efficient cross-agent communication and memory could support larger groups. Combined with efficient video generation, these extensions could enable real-time, interactive multi-agent world simulation.
We hope this work advances world models toward simulating multiple embodied agents in a persistent and evolving world.

\bibliography{iclr2027_conference}
\bibliographystyle{iclr2027_conference}

\appendix
\newpage
\section*{\Large Appendix}
{
    \renewcommand{\baselinestretch}{1.8}\selectfont

    \noindent\textbf{A\quad Additional Results}
    
    \hspace*{2em}A.1\quad Additional Qualitative Results
    \vspace{-5pt}
    
    \hspace*{2em}A.2\quad Generalization to Extended Settings

    \noindent\textbf{B\quad Evaluation Details}

    \hspace*{2em}B.1\quad Shared-World Consistency Metrics
    \vspace{-5pt}
   
    \hspace*{2em}B.2\quad Evaluation Metrics
    \vspace{-5pt}
    
    \hspace*{2em}B.3\quad Baseline Implementation Details
    
    \noindent\textbf{C\quad Dataset Construction}
    
    \hspace*{2em}C.1\quad Synthetic Dataset Generation
    \vspace{-5pt}
    
    \hspace*{2em}C.2\quad Real Dataset Curation
    \vspace{-5pt}
    
    \hspace*{2em}C.3\quad Benchmark Construction

    \noindent\textbf{D\quad Implementation Details}
    
    \hspace*{2em}D.1\quad Model Details
    \vspace{-5pt}   
    
    \hspace*{2em}D.2\quad Training Details

    \vspace{10pt}
}

\vfill

\clearpage
\section{Additional Results}

\subsection{Additional Qualitative Results}
\label{sec:supp_additional_qualitative}
\begin{figure}[t!]       
    \centering    
    \includegraphics[width=\linewidth]{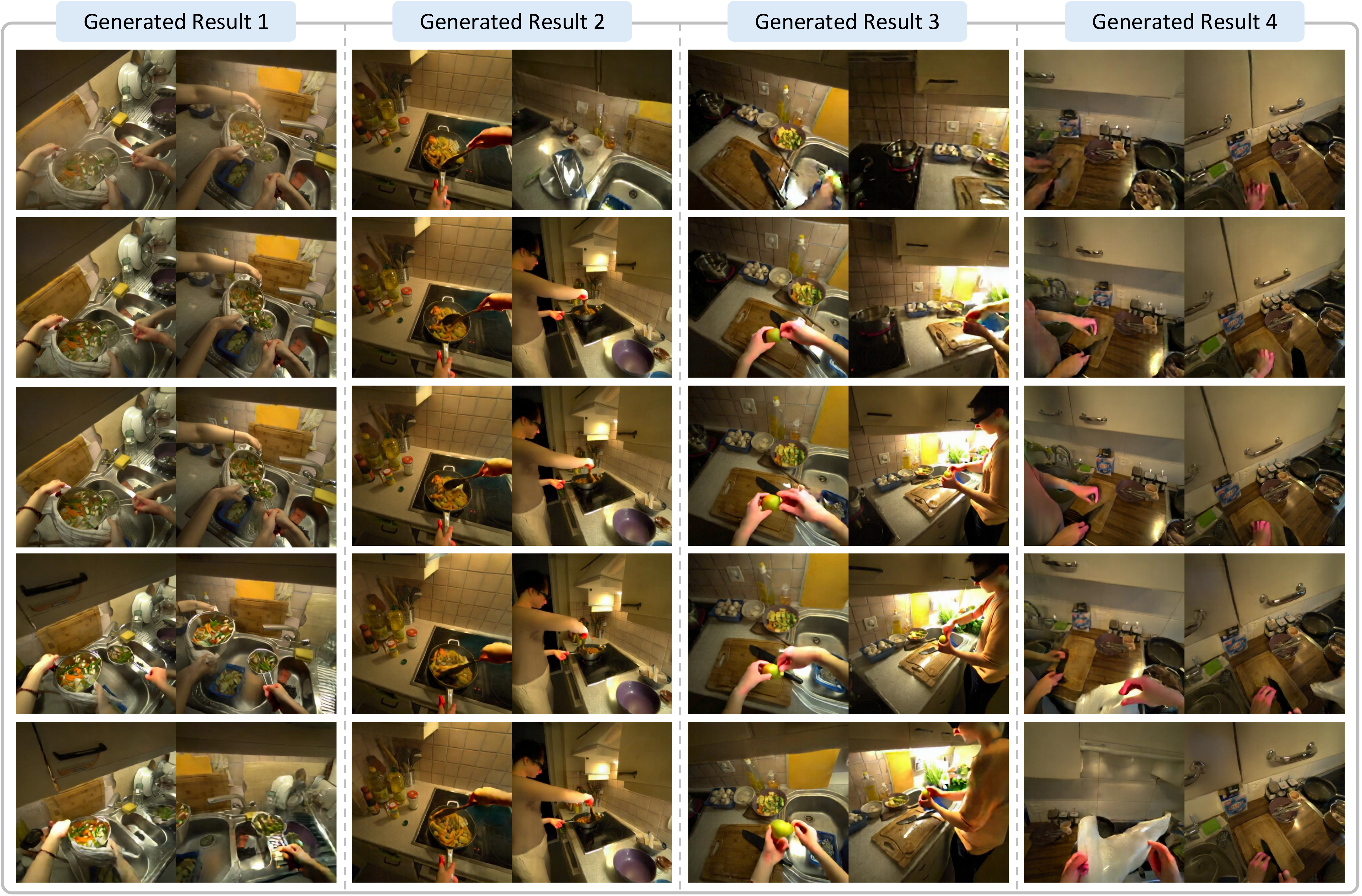}
    \vspace{-20pt}
    \caption{
\textbf{Additional Qualitative Results on the Real Benchmark.}
We show additional generations from diverse interaction sequences in real benchmark.
Each example contains the two synchronized ego streams generated jointly by our model across multiple timesteps. Best viewed zoomed in.}
    \vspace{-15pt}
    \label{fig:supp_qual_real}   
\end{figure}

\begin{figure}[t!]       
    \centering    
    \includegraphics[width=\linewidth]{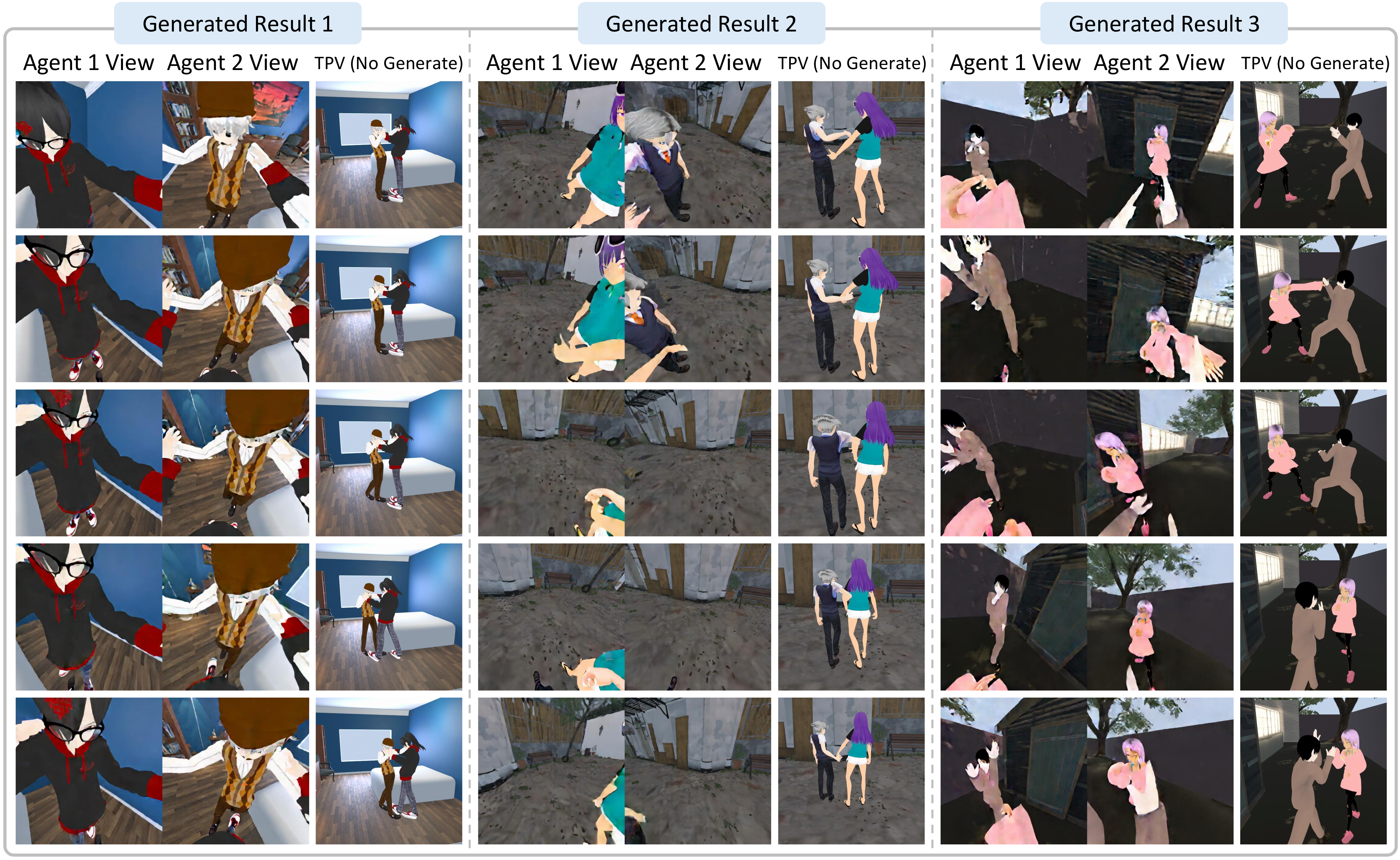}
    \vspace{-20pt}
\caption{
\textbf{Additional Qualitative Results on the Synthetic Benchmark.}
We show additional generations from diverse interaction sequences in synthetic benchmark.
For each example, the two ego views are jointly generated by our model.
The third-person view (TPV) is shown only as a visual reference to help interpret the shared scene and interaction, and is not generated or used as input by our model. Best viewed zoomed in.
}
    \vspace{-15pt}
    \label{fig:supp_qual_synth}   
\end{figure}

We provide additional qualitative results on the real and synthetic benchmarks in Fig.~\ref{fig:supp_qual_real} and~\ref{fig:supp_qual_synth}.
Across diverse real-world interactions, our model generates coherent paired ego views that preserve shared scene content and compatible interaction dynamics despite substantial viewpoint differences between the two agents.
On the synthetic benchmark, where agent identity and motion can be inspected more clearly, the model consistently preserves the appearance of both agents and their coordinated interactions across the generated ego views.

\subsection{Generalization to Extended Settings}
\label{sec:supp_appqual}
\begin{figure}[t!]       
    \centering    
    \includegraphics[width=\linewidth]{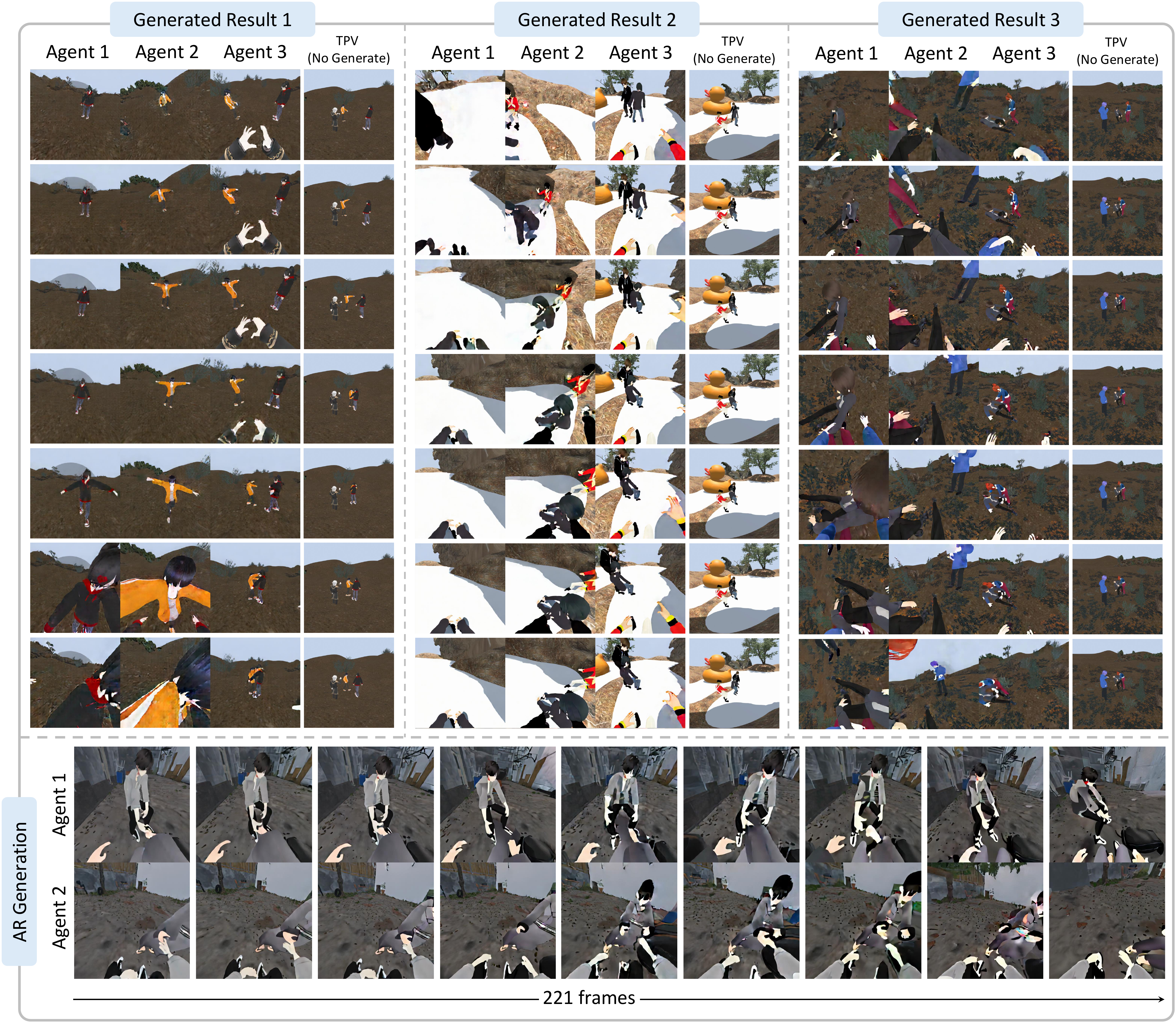}
    \vspace{-20pt}
\caption{
\textbf{Multi-Agent and Autoregressive Generation.}
We further demonstrate our model in more diverse generation settings, including multi-agent and long-horizon generation.
Our model jointly generates three synchronized ego views for three interacting agents (top), and autoregressively generates two-agent sequences of 221 frames (bottom).
For the three-agent results, the third-person view (TPV) is provided only as a visual reference to help interpret the shared scene and interaction, and is not generated or used as input by our model. Best viewed zoomed in.
}
    \vspace{-15pt}
    \label{fig:supp_appqual_synth}   
\end{figure}

We further demonstrate our model in diverse generation settings beyond the standard two-agent, fixed-length setting used for training.

\paragrapht{Multi-Agent Generation.}
Our framework extends naturally to a larger number of agents without modifying the underlying architecture. To add a third agent, we simply append an additional ego stream and its corresponding agent token, project the third agent's body and hand poses into each ego view in the same manner as in the two-agent setting, and include the resulting conditions in joint generation. All ego streams are then processed together with shared attention and the same shared environment memory. As shown in Fig.~\ref{fig:supp_appqual_synth} (top), this simple extension enables the model to jointly generate three synchronized ego views for three interacting agents while preserving a common shared scene and their mutual interactions.

\paragrapht{Autoregressive Generation.}
We further extend generation along the temporal dimension by autoregressively chaining multiple generation chunks. As shown in Fig.~\ref{fig:supp_appqual_synth} (bottom), this produces a 221-frame two-agent sequence while maintaining the interacting agents and shared environment over the extended horizon. Each subsequent chunk is conditioned on two frames from the preceding generated chunk, while generation within each chunk remains bidirectional. To improve robustness to imperfect generated context, we perturb the conditional latents during training. We randomly condition on one or two frames, and when two frames are used, independent Gaussian noise with standard deviation $0.1$ is applied with probability $0.5$. The denoising target remains clean, exposing the model to imperfect handoff frames without requiring autoregressive rollouts during training.

\section{Evaluation Details}

\subsection{Shared-World Consistency Metrics}
\label{sec:supp_shared_world_metric_main}
\subsubsection{Metric Implementation Details}
\label{sec:supp_shared_world_metric}

We provide implementation details for the shared-world consistency metrics
introduced in Sec.~\ref{sec:shared_world_metric}.
Following the notation in the main paper, we consider an ordered pair of
distinct agents $(n,m)$, where agent $n$ is the query stream and agent $m$
is the corresponding target stream.
For clarity, we describe the metrics for a synchronized frame and omit the
frame index $t$.
We report $S_{\mathrm{env}}$, $S_{\mathrm{update}}$ on both benchmarks, and $S_{\mathrm{id}}$ on synthetic.

\paragrapht{Feature Representation and Change Mask.}
We extract DINOv3~\citep{simeoni2025dinov3} ViT-L/16 features on a
$32\times32$ patch grid, denoted by $\phi(\cdot)$.
For each patch $p$ in the generated query view $X_{\mathrm{gen}}^{(n)}$,
we compute its discrepancy from the stream-aligned warped condition:
\begin{equation}
d_{\mathrm{gen}}^{(n)}(p)
=
\left\|
\phi(X_{\mathrm{gen}}^{(n)})(p)
-
\phi(X_{\mathrm{warp}}^{(n)})(p)
\right\|_2.
\end{equation}

To account for reprojection artifacts, we set the change threshold
$\tau^{(n)}$ to the 80th percentile of feature distances over valid
non-human background patches in each frame.
The generated change mask is then
\begin{equation}
M_{\mathrm{update}}^{(n)}(p)
=
\mathbbm{1}
\left[
d_{\mathrm{gen}}^{(n)}(p)>\tau^{(n)}
\right].
\end{equation}
All visibility and change masks are evaluated on the same patch grid.

\paragrapht{Cross-View Correspondence.}
As illustrated in Fig.~\ref{fig:eval_protocol} (a), we establish
correspondences between the synchronized ground-truth views
$X_{\mathrm{gt}}^{(n)}$ and $X_{\mathrm{gt}}^{(m)}$ using metric depth
and ground-truth camera poses.
Each patch center $p$ in the query view is back-projected into 3D and
projected into the target view to obtain its corresponding patch $q$.
We retain correspondences whose projected locations lie within the image
boundaries and whose reprojected depths agree with the target-view depths
within $10$ cm.
The resulting co-visible correspondence set is denoted by
$\mathcal{C}^{(n\rightarrow m)}$.

For each $(p,q)\in\mathcal{C}^{(n\rightarrow m)}$, we compute the
cross-view feature similarity between the corresponding generated regions:
\begin{equation}
s^{(n\rightarrow m)}(p,q)
=
\operatorname{cos}
\left(
\phi(X_{\mathrm{gen}}^{(n)})(p),
\phi(X_{\mathrm{gen}}^{(m)})(q)
\right).
\end{equation}

\paragrapht{Environment Consistency.}
$S_{\mathrm{env}}$ measures whether environment content generated without
direct scene evidence remains consistent across ego streams.
As illustrated in Fig.~\ref{fig:eval_protocol} (b), we select co-visible
correspondences whose query patch is not covered by the stream-aligned
geometric memory:
\begin{equation}
\mathcal{C}_{\mathrm{env}}^{(n\rightarrow m)}
=
\left\{
(p,q)\in\mathcal{C}^{(n\rightarrow m)}
\;\middle|\;
M_{\mathrm{vis}}^{(n)}(p)=0
\right\}.
\end{equation}
For our two-agent setting, we average both ordered directions:
\begin{equation}
S_{\mathrm{env}}
=
\frac{1}{2}
\sum_{(n,m)\in\{(1,2),(2,1)\}}
\operatorname{Avg}_{(p,q)\in
\mathcal{C}_{\mathrm{env}}^{(n\rightarrow m)}}
s^{(n\rightarrow m)}(p,q).
\end{equation}
The visibility constraint is applied only to the query patch $p$; its
corresponding target patch $q$ need not be uncovered.

\paragrapht{State Update Consistency.}
$S_{\mathrm{update}}$ measures whether interaction-induced state changes
are represented consistently across ego streams.
As illustrated in Fig.~\ref{fig:eval_protocol} (c), we select co-visible
correspondences whose query patch differs from valid warped scene evidence:
\begin{equation}
\mathcal{C}_{\mathrm{update}}^{(n\rightarrow m)}
=
\left\{
(p,q)\in\mathcal{C}^{(n\rightarrow m)}
\;\middle|\;
M_{\mathrm{update}}^{(n)}(p)=1,\;
M_{\mathrm{vis}}^{(n)}(p)=1
\right\}.
\end{equation}
We then compute
\begin{equation}
S_{\mathrm{update}}
=
\frac{1}{2}
\sum_{(n,m)\in\{(1,2),(2,1)\}}
\operatorname{Avg}_{(p,q)\in
\mathcal{C}_{\mathrm{update}}^{(n\rightarrow m)}}
s^{(n\rightarrow m)}(p,q).
\end{equation}
Requiring $M_{\mathrm{vis}}^{(n)}(p)=1$ ensures that the detected change is
measured relative to valid scene appearance evidence rather than an
uncovered region.

\paragrapht{Identity Consistency.}
On synthetic, $S_{\mathrm{id}}$ evaluates whether the generated interaction
partner preserves its intended avatar appearance.
As illustrated in Fig.~\ref{fig:eval_protocol} (d), for each benchmark avatar, we construct a multi-view reference images by rendering the
avatar from 12 azimuths under full-body and upper-body framings and encode
the masked renders using the CLS token of DINOv3 ViT-L/16~\citep{simeoni2025dinov3}.
For each generated frame after the provided first frame, we detect the
interaction partner using SAM3~\citep{carion2026sam} and encode the detected
character using the same feature extractor.
We compare its feature against the multi-view reference images of the
corresponding ground-truth avatar and take the maximum cosine similarity
over the reference views.
$S_{\mathrm{id}}$ is obtained by averaging these similarities over all
valid frames and both ego streams.

\subsubsection{Metric Validation}
\label{sec:supp_metric_validation}
\begin{figure}[t!]       
    \centering    
    \includegraphics[width=\linewidth]{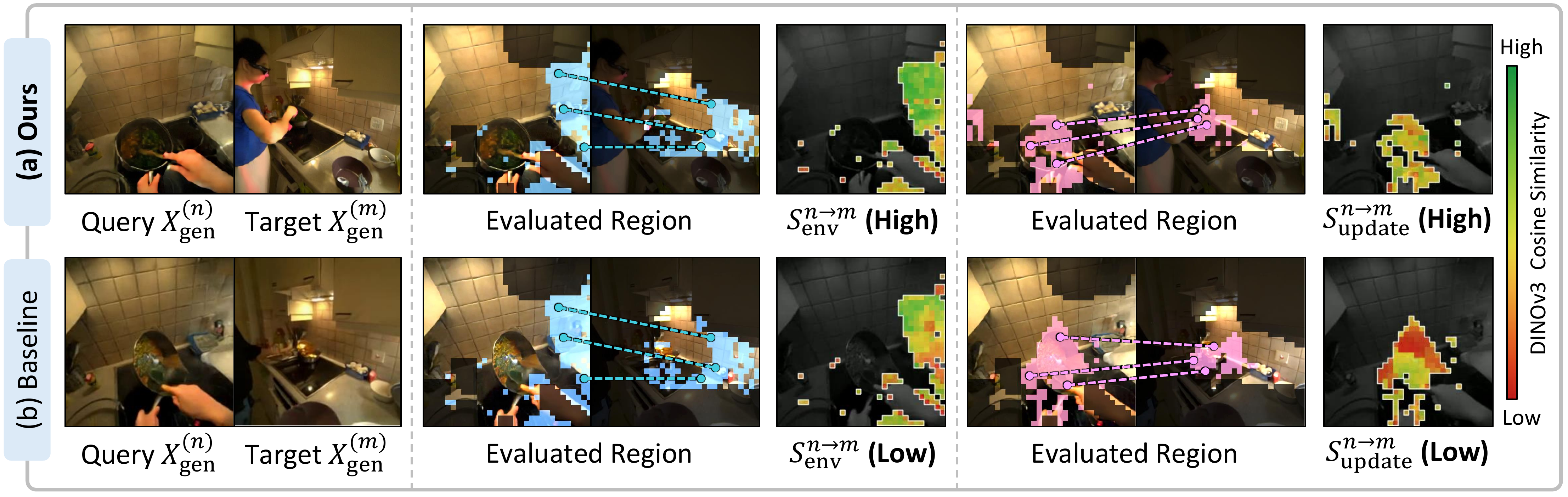}
    \vspace{-20pt}
\caption{
\textbf{Metric validation.}
We visualize the query regions evaluated by $S_{\mathrm{env}}$ and
$S_{\mathrm{update}}$, their co-visible correspondences in the target
stream, and the resulting DINOv3 cosine similarities.
The consistent generation in (a) shows high correspondence similarity and
receives higher metric scores, whereas the cross-view inconsistencies in
(b) appear as localized similarity drops and lower scores.
This correspondence provides qualitative evidence that the proposed metrics
capture the intended cross-view consistency. Best viewed zoomed in.
}
    \vspace{-15pt}
    \label{fig:supp_metric_validation}   
\end{figure}

\paragrapht{Qualitative Validation.}
Fig.~\ref{fig:supp_metric_validation} visualizes how the proposed metrics
respond to actual generation successes and failures.
We highlight the query regions selected for $S_{\mathrm{env}}$ and
$S_{\mathrm{update}}$ and visualize the DINOv3 cosine similarity of their
co-visible correspondences in the other ego stream.
Our generation preserves high similarity in both regions, yielding
$S_{\mathrm{env}}=0.582$ and $S_{\mathrm{update}}=0.495$.
In contrast, independent single-stream generation shows visible
cross-view disagreement in the same regions, reflected by lower similarities
and scores of $0.498$ and $0.402$, respectively.
For clarity, the figure visualizes one correspondence direction, while the
reported scores use bidirectional evaluation.

\paragrapht{Controlled Validation.}
We further validate the metric behavior by introducing controlled
cross-view inconsistencies into 100 benchmark clips
(Tab.~\ref{tab:metric_robustness}).
Only one ego stream is modified and the paired stream is kept unchanged,
while the metrics are computed with the standard bidirectional evaluation.
We consider three cases: temporal misalignment by shifting one stream
by five frames, an environment conflict by replacing content without direct
scene evidence ($M_{\mathrm{vis}}=0$) with a fixed appearance, and a state
conflict by similarly replacing regions containing interaction-induced
state changes ($M_{\mathrm{gen}}=1, M_{\mathrm{vis}}=1$).
\begin{wraptable}{r}{0.5\linewidth}
\vspace{-6pt}
\centering
\caption{
\textbf{Controlled validation of the proposed metrics.}
We evaluate how $S_{\mathrm{env}}$ and $S_{\mathrm{update}}$ respond to
temporal misalignment, environment conflict, and state conflict.
Values in brackets denote changes from the original generations.
}
\label{tab:metric_robustness}
\resizebox{\linewidth}{!}{
\begin{tabular}{l|c|ccc}
\toprule
\textbf{Metric}
& \textbf{Ours}
& \textbf{Misalignment}
& \textbf{Env. Conflict}
& \textbf{State Conflict} \\
\midrule
$S_{\mathrm{env}}$
& 0.468
& 0.399 {\scriptsize($-0.069$)}
& \textbf{0.377} {\scriptsize(\boldsymbol{$-0.090$})}
& 0.408 {\scriptsize($-0.059$)} \\
$S_{\mathrm{update}}$
& 0.466
& 0.377 {\scriptsize($-0.090$)}
& 0.447 {\scriptsize($-0.019$)}
& \textbf{0.320} {\scriptsize(\boldsymbol{$-0.147$})} \\
\bottomrule
\end{tabular}
}
\vspace{-8pt}
\end{wraptable}

Temporal misalignment decreases both metrics.
More specifically, the environment conflict decreases
$S_{\mathrm{env}}$ by $0.090$ while changing $S_{\mathrm{update}}$ by only
$0.019$, whereas the state conflict decreases $S_{\mathrm{update}}$ by
$0.147$, compared with $0.059$ for $S_{\mathrm{env}}$.
These differentiated responses support that the two metrics capture their
intended, complementary aspects of shared-world consistency.

\subsection{Evaluation Metrics}
\label{sec:supp_eval_metrics}
In addition to shared-world consistency, we evaluate
camera control, self-action control, other-agent action
control, and video quality.

\paragrapht{Camera Control.}
We estimate camera trajectories from generated videos using
DA3-GIANT~\citep{lin2025depth} and align them to the ground-truth
trajectories with a similarity transformation to resolve
monocular scale ambiguity.
We report the absolute trajectory error (ATE) of camera
centers and the mean rotation error between aligned
and ground-truth camera orientations.

\paragrapht{Action Control.}
Following E$^3$C~\citep{gu20263}, we evaluate
self-action and other-agent action control separately.
For self-action control, we segment the wearer's hands
using SAM3~\citep{carion2026sam} and report Hand-F1 and Hand-mIoU
for hand presence and spatial alignment, respectively.
For other-agent action control, we detect the interaction
partner and estimate body keypoints using YOLOX~\citep{ge2021yolox}
and ViTPose~\citep{xu2022vitpose}.
We report Full-Body-F1 for partner detection and Full-Body-PCK@10
for pose alignment over matched detections.

\paragrapht{Video Quality.}
We report PSNR~\citep{huynh2008psnr}, SSIM~\citep{wang2004ssim}, and LPIPS~\citep{zhang2018lpips} between generated and
ground-truth videos to measure reconstruction fidelity
and perceptual similarity.
We additionally compute FVD~\citep{unterthiner2019fvd} using I3D~\citep{carreira2017i3d} features to
assess video-level distributional similarity.

\subsection{Baseline Implementation Details}
\label{sec:supp_baseline_details}

\subsubsection{Multi-View Video Generation}
\paragrapht{GEN3C~\citep{ren2025gen3c}.}
We use the GEN3C-Cosmos-7B checkpoint
($704\times1280$, 121-frame chunks).
For each generation direction, we construct a per-frame
4D cache from the source view's ground-truth RGB video,
metric depth estimated by DA3~\citep{lin2025depth},
and ground-truth camera parameters.
We render the cache along the target camera trajectory,
using temporally corresponding source frames,
and provide the rendered observations and validity
masks as warp conditions, together with the target
initial frame and caption.
We letterbox the $504\times504$ inputs into
$704\times1280$, adjusting intrinsics accordingly.
We pad the 77-frame camera trajectory to 121 frames
by repeating its final pose and truncate the output
to 77 frames.
We use 35 denoising steps, guidance scale 1.0,
and point-filter threshold 0.05.
The central $704\times704$ region is cropped
before evaluation.

\paragrapht{AnyView-DVS~\citep{van2026anyview}.}
We use the AnyView-DVS-2B checkpoint, conditioned
on the source ground-truth video and both cameras'
intrinsics and extrinsics through Pl\"ucker rays.
The target initial frame, text, and depth are not provided.
We re-anchor poses to the target camera at frame 0
and scale translations by $1/8$ following the
released configuration.
We resize the inputs to $576\times576$ with
adjusted intrinsics and generate 41 frames,
the maximum supported length ($1+4k$).
We use 35 sampling steps and guidance scale 0.

\subsubsection{Single-Ego World Models}

\paragrapht{EgoSim~\citep{hao2026egosim}.}
We use EgoSim-14B based on Wan2.1-Fun-14B-InP.
Following the released pipeline, we construct
an environment prior by back-projecting each view's
ground-truth initial frame using estimated metric depth
and rendering it along the ground-truth camera trajectory,
together with a coverage mask.
We render the wearer's Aria MPS hand tracking
(21 landmarks per hand) into a 3D skeleton video
using the official rendering parameters.
We additionally provide the initial-frame CLIP
embedding and corresponding caption.
Inputs are letterboxed into $480\times832$ with
adjusted intrinsics.
We generate 61 frames using the first 61 frames
of the conditions, retaining the benchmark's
10 FPS indexing rather than resampling to the
model's nominal 16 FPS.
We use 50 denoising steps without classifier-free guidance.

\paragrapht{JointControlVideo~\citep{zhang2026jointcontrolvideo}.}
We use the released LoRA and hand-controller weights.
Each view is independently conditioned on its
ground-truth initial frame, caption, and the wearer's
3D hand-joint trajectories.
We reorder Aria MPS hand landmarks into the
MANO-21 joint ordering and provide 42 joints
in the camera coordinate system, along with
camera intrinsics.
Untracked hands are zeroed, consistent with
training-time joint dropout.
We letterbox inputs into $480\times832$ and
generate 77 frames using 50 denoising steps
and guidance scale 5.0.
The model does not explicitly receive camera trajectories.

\subsubsection{General World Models}
\paragrapht{DreamX-World~\citep{team2026dreamx}.}
We use the DreamX-World-5B-Cam checkpoint based on
Wan2.2-TI2V-5B.
Each ego stream is independently generated from its
ground-truth initial frame, caption, and camera trajectory.
We resize the initial frame to $480\times480$ using
bicubic interpolation and generate 77 frames at the
same resolution.
The official inference interface converts discrete
camera commands into per-frame poses.
We bypass this conversion and directly provide
the recorded camera poses in the same internal format.
The model interpolates the poses to 20 latent frames
and expresses them relative to the initial camera,
using fixed intrinsics and metric-scale translations
without normalization.

\paragrapht{LingBot-World~\citep{team2026advancing}.}
We use the LingBot-World-Base-Cam checkpoint based on
Wan2.2-I2V-A14B.
Each ego stream is independently generated from its
ground-truth initial frame, caption, and camera trajectory.
We resize the initial frame to $480\times480$ using
bicubic interpolation and generate 77 frames at the
same resolution.
We provide per-frame camera-to-world poses and intrinsics.
Following the official implementation, the camera poses
are interpolated to 20 latent frames, expressed relative
to the initial camera, and normalized by the maximum
translation magnitude before Pl\"ucker-ray encoding.
We compensate for the repository's intrinsic rescaling
from its $480\times832$ reference canvas to preserve
the effective intrinsics at $480\times480$.

\subsubsection{Multi-Agent World Models}
\label{sec:supp_adaptation_details}

\paragrapht{MultiWorld~\citep{wu2026multiworld}.}
We adapt MultiWorld to our embodied multi-agent setting using the same
video backbone, training data, and Shared Action Conditioning as \ours.
We retain our Joint Multi-Agent Generation, jointly denoising the two
ego streams through full self-attention over the concatenated stream tokens.
For environment conditioning, we replace our Shared Environment Memory
with MultiWorld's VGGT-based Global State Encoder.
We extract a shared global representation from the observation history
using frozen VGGT features and inject the resulting representation into
the ego streams through cross-attention.
Our stream-aligned geometric memory and cross-stream anchor memory are
not used in this adaptation.

\paragrapht{Solaris~\citep{savva2026solaris}.}
We adapt Solaris using the same backbone, training data, and
Shared Action Conditioning as \ours.
The two ego streams are jointly denoised using Joint Multi-Agent Generation,
allowing their tokens to interact through full self-attention.
Unlike \ours, this adaptation does not use an additional explicit
environment-conditioning module.
It therefore corresponds to joint multi-agent generation with
Shared Action Conditioning but without Shared Environment Memory.

\paragrapht{$\gamma$-World~\citep{liu2026gamma}.}
We adapt $\gamma$-World using the same backbone, training data, and
Shared Action Conditioning as \ours.
Instead of our Joint Multi-Agent Generation, we retain
$\gamma$-World's multi-agent interaction mechanism based on
Sparse Hub Attention and Simplex Rotary Agent Encoding.
The ego streams communicate through the shared hub tokens rather than
full self-attention over a single concatenated sequence.
We do not introduce an additional explicit environment-conditioning
module, isolating the effect of its native cross-agent communication
mechanism under the same embodied action conditions.

\paragrapht{MetaWorld~\citep{hu2026metaworld}.}
We adapt MetaWorld using the same backbone, training data, and
Shared Action Conditioning as \ours.
Rather than jointly processing all ego tokens with full self-attention,
we retain MetaWorld's World-State Alignment mechanism.
Since no official implementation is publicly available, we reimplement
this module based on the method described in the paper, where the ego
branches are processed separately and exchange synchronized information
through cross-view attention.
For environment conditioning, we provide geometric scene evidence using
warped observations and depth-based conditions, following the MetaWorld
formulation.

\section{Dataset Construction}
\label{sec:data_construction}
\begin{figure}[t!]       
    \centering    
    \includegraphics[width=\linewidth]{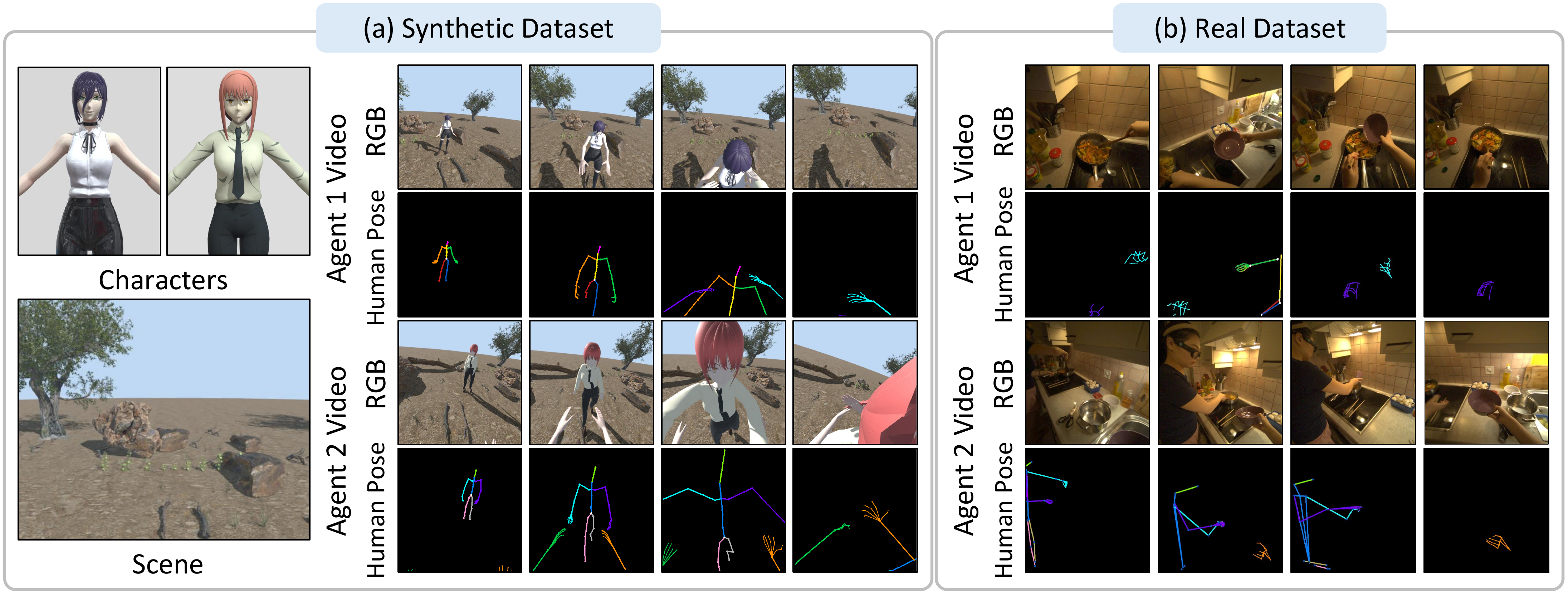}\
    \vspace{-20pt}
    \caption{\textbf{Dataset example.} Examples are from (a) our generated synthetic dataset and (b) curated real-world CoMind dataset, showing synchronized ego streams and corresponding human poses.}
        \vspace{-10pt}
    \label{fig:supp_dataset_example}   
\end{figure}

We construct training data from both synthetic multi-person interactions and
real-world synchronized egocentric recordings, as illustrated in
Fig.~\ref{fig:supp_dataset_example}.
The synthetic pipeline provides diverse agent appearances, environments, and
human-human interactions with complete geometric and motion annotations,
whereas the real-world data captures natural interactions and visual complexity.
For both sources, we process paired ego streams into a common training format
consisting of scene, pose, camera, reference-image, and text conditions.

\subsection{Synthetic Dataset Generation}
\label{sec:supp_synthetic_dataset}
\begin{figure}[t!]       
    \centering    
    \includegraphics[width=\linewidth]{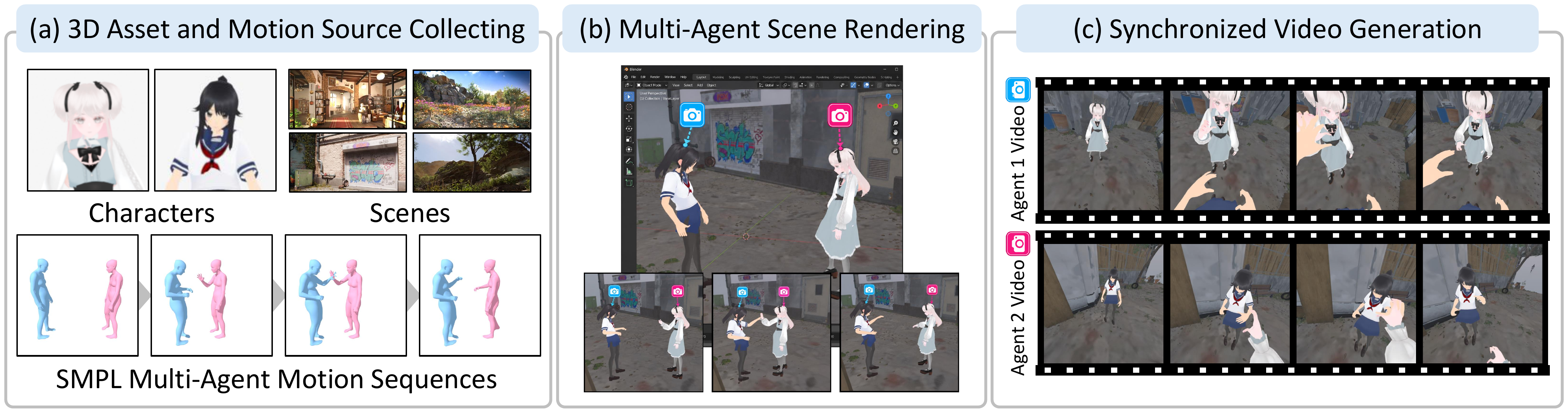}
    \vspace{-20pt}
    \caption{\textbf{Synthetic dataset generation pipeline.} (a) We collect diverse 3D characters, environments, and paired motions, (b) apply the motions to characters, place them in the shared world space, and attach egocentric cameras, and (c) render synchronized egocentric video pairs.}
    \vspace{-15pt}
    \label{fig:supp_synthetic_construction}   
\end{figure}

To complement the limited diversity of real data, we construct synchronized
egocentric video pairs from interacting avatars in shared 3D environments, as illustrated in Fig.~\ref{fig:supp_dataset_example} (a).
As illustrated in Fig.~\ref{fig:supp_synthetic_construction}, paired human motions
are retargeted to rigged avatars, placed into diverse scenes, and rendered
synchronously from two head-mounted cameras.
The rendered geometry and poses allow us to construct the same conditioning
modalities used for real data, together with shared posed reference images.

\paragrapht{Assets and Motion Retargeting.}
We use 1,822 two-person interaction sequences from
Inter-X~\citep{xu2024inter} (1,372 sequences across 40 categories)
and InterHuman~\citep{liang2024intergen} (450 sequences),
resampled to 10 FPS.
We collect 100 rigged VRM avatars and seven indoor and outdoor Blender
environments.
For each sequence, we randomly assign two avatars and transfer the SMPL-X
motion to their skeletons.
Because avatar body proportions can differ substantially from the source
motions, we adapt the retargeted arm motion according to avatar limb lengths.
Analytic two-bone inverse kinematics further aligns the wrists with their
source trajectories, reducing contact errors in interactions such as
handshakes.
Finger articulation is transferred separately using relative joint flexion,
and the retargeted pair is vertically aligned to the scene floor.

\paragrapht{Scene Placement and Rendering.}
We define scene anchors with valid standing positions and orientations and
transform each interaction into a canonical pair-centric coordinate system
before placement.
Ray casting estimates available free space around each anchor, allowing motion
trajectories to be assigned only to geometrically compatible locations.
We discard placements with frequent camera--geometry intersections, yielding
53 validated scene configurations across the seven environments.
For compatible placements, we additionally favor configurations with sufficient
mutual view coverage between the two agents, which provides useful shared
visual evidence for multi-ego generation.

We attach a head-mounted camera to each avatar and synchronously render the two
egocentric streams with a $100^\circ$ horizontal field of view at
$504\times504$ resolution and 10 FPS using Blender EEVEE.
The wearer's head geometry is hidden from its own camera to avoid near-camera
artifacts.
RGB images, metric depth, and person-instance masks are rendered jointly.
Avatar animations and cameras are keyframed over the complete interaction to
preserve exact temporal synchronization between the two ego streams.

\paragrapht{Condition Extraction.}
We divide the rendered sequences into non-overlapping 77-frame windows and
construct the conditioning modalities used by our model.
For explicit scene conditioning, we select observations from the available
history according to their target-view reprojection coverage.
Human pixels are removed using the rendered instance masks before
back-projecting the remaining RGB-D observations into 3D and reprojecting them
into each target camera.
This produces the warped scene observation and its corresponding visibility
mask.

For shared action conditioning, we project the articulated skeletons of both
agents into each target ego view.
Body and finger joints are rendered using consistent identity-specific colors,
allowing the model to distinguish the same two agents across viewpoints.
Camera motion is represented using per-pixel Pl\"ucker ray maps constructed in
the shared coordinate system.

We additionally construct the implicit shared environment memory by selecting
eight clean reference observations shared across the two streams.
References are greedily selected to maximize complementary scene coverage over
both target ego views.
For each reference, we retain its RGB image, camera condition, and
identity-colored pose condition.
Because the underlying interaction is already specified by the pose
conditioning, we use a generic first-person interaction prompt for synthetic
training samples.

\paragrapht{Quality Control and Dataset Statistics.}
We filter sequences with invalid viewing conditions, prolonged extreme partner
proximity, or severe camera--geometry intersections.
A wall-intrusion score based on near-camera scene geometry provides an
additional check for cases in which the virtual head-mounted camera enters
scene surfaces.
The resulting synthetic corpus contains 4,736 paired 77-frame clips,
corresponding to 9,472 egocentric videos, generated from 1,822 interaction
sequences, 100 avatars, seven environments, and 53 validated scene
configurations.
All videos are rendered at 10 FPS and $504\times504$ resolution.

\subsection{Real Dataset Curation}
\label{sec:supp_real_dataset}

We curate CoMind~\citep{gavryushin2026comind}, which records two interacting
wearers during collaborative kitchen activities, into synchronized two-ego
training pairs, as illustrated in Fig.~\ref{fig:supp_dataset_example} (b).
The fisheye camera streams are rectified to a pinhole projection and processed
at 10 FPS and $504\times504$ resolution.
Each training example contains 200 context frames followed by 77 target frames.

\paragrapht{Geometry and Scene Conditioning.}
We use CoMind's multi-device SLAM reconstruction for recordings with a valid
shared solve, which places both head-mounted cameras in a common coordinate
system.
For every ego pair, we express both cameras in a shared canonical coordinate
frame and convert their intrinsics and extrinsics into per-pixel Pl\"ucker ray
maps, preserving the relative camera geometry between the two views.

Because the released scene geometry is too sparse to directly construct dense
target-view appearance conditions, we estimate per-frame metric depth using
DA3~\citep{lin2025depth} with the corresponding camera parameters.
For each target view, we back-project valid pixels from past observations into
3D and reproject them using the target camera to construct the warped scene
appearance condition.
The valid reprojection region defines the corresponding binary visibility
mask.

Moving people must not be treated as part of the persistent scene memory.
We therefore segment human regions using SAM3~\citep{carion2026sam} and remove
them from source observations before warping.
For each target frame, we select the source observation with the largest valid
reprojection coverage after human masking.
This provides an explicit geometric memory of previously observed static scene
content while preventing ghosting or leakage from past human locations.

\paragrapht{Pose Conditioning.}
CoMind provides egocentric hand poses but does not contain complete whole-body
motion annotations.
We therefore estimate full-body 2D poses using
Sapiens2~\citep{khirodkar2026sapiens2}, including body, hand, and facial
landmarks.
We associate detected people across the synchronized ego streams and maintain
consistent identities over time.
The two agents are rendered with fixed identity-specific colors across both
views so that each ego stream receives the projected body motion of both the
wearer and the interacting partner in a shared identity convention.

\paragrapht{Cross-Stream Anchor Memory Construction.}
In addition to stream-aligned geometric memory, we construct clean reference
observations that provide complementary appearance information about the shared
environment.
Candidate observations are selected from the available context of both ego
streams.
Following the synthetic pipeline, we greedily select references according to
their complementary target-view coverage and retain their RGB appearance,
camera conditions, and identity-consistent pose conditions.
The resulting references are shared by both generated streams and form the
implicit reference memory used by our model.

\paragrapht{Text Conditioning.}
CoMind does not provide textual descriptions for the egocentric videos.
We therefore generate a description for each ego stream using
Qwen3.6-35B-A3B~\citep{qwen36_35b_a3b}.
The captions describe the visible activity and interaction from the
corresponding viewpoint and are used as per-view text conditions during
training.

\paragrapht{Quality Control and Dataset Statistics.}
We retain recordings for which the multi-device SLAM reconstruction provides a
reliable shared camera solution and remove windows with unreliable camera
registrations or invalid visual conditions.
After filtering, the real training corpus contains approximately 12K
synchronized two-ego clip pairs from 32 recordings.
Each pair is processed at 10 FPS and $504\times504$ resolution with 200 context
frames and 77 target frames.

\subsection{Benchmark Construction}
\label{sec:supp_benchmark_construction}

\paragrapht{Real Benchmark.}
We sample 100 clips from the official validation split of
CoMind~\citep{gavryushin2026comind}.
These recordings are disjoint from our training data and contain unseen
kitchens and participants.
We process them using the same geometry, pose, and camera pipeline described
above and extract synchronized 77-frame windows from the two ego streams.
All metrics are computed at $480\times480$ resolution.
We exclude clips with hair occluding the ego camera, mean cross-view
co-visibility below 0.075, or fewer than 10 partner-visible frames in both
views.
Co-visibility is computed through cross-view depth reprojection using the
recorded cameras with a 10\,cm depth-consistency tolerance.
To avoid concentrating the benchmark on a narrow range of viewpoint overlap,
we allocate recording-level quotas proportional to each recording's eligible
candidate pool, yielding 28, 25, and 47 clips.
Within each recording, candidates are sorted by co-visibility, divided into
equal-count strata, and randomly sampled across these strata.
The resulting benchmark has a mean cross-view co-visibility of 0.38 and a
median of 43 partner-visible frames per clip.

\paragrapht{Synthetic Benchmark.}
We construct 100 evaluation clips using the same rendering and conditioning
pipeline as our synthetic training data while using held-out assets.
The benchmark spans six Poly Haven~\citep{polyhaven_scenes} environments:
Hidden Alley, Verdant Trail, The Shed, Namaqualand, Pawn Shop, and Pine Forest.
Candidate windows must contain at least 10 partner-visible frames with at
least six body joints inside the image, and must have mean cross-view
co-visibility of at least 0.05.
We retain at most one window from each interaction sequence to avoid
near-duplicate samples.
Scene-level quotas are assigned proportional to the eligible validation pool,
resulting in 14, 23, 18, 14, 19, and 12 clips for the six scenes,
respectively.
Within each scene, we perform stratified sampling by co-visibility following
the real benchmark protocol.
The resulting synthetic benchmark contains 100 distinct interaction sequences,
including 84 from Inter-X and 16 from InterHuman, across 86 avatar pairs.
It has a mean cross-view co-visibility of 0.31 and provides exact camera,
depth, segmentation, pose, and identity annotations for controlled evaluation.

\section{Implementation Details}
\subsection{Model Details}
\label{sec:supp_model_details}

\paragrapht{Model Configuration.}
We initialize \ours from the Cosmos-Predict2.5 2B\citep{ali2025world}
multi-view diffusion transformer, with a frozen Wan-2.1
causal video VAE and Cosmos-Reason1 7B text encoder.
Each training sample contains two synchronized 77-frame
ego streams at $480\times480$ resolution and 10 FPS,
encoded into 20 latent frames per view.
We use first-frame conditioning and four shared
references selected by greedy coverage maximization
over both views.

\subsection{Training Details}
\label{sec:supp_training_details}
\paragrapht{Training.}
We train separate checkpoints for Real and Synthetic,
using CoMind and our synthetic dataset, respectively,
without transferring weights between them.
For each model, we fine-tune the DiT and conditioning
modules using the rectified-flow objective, applying
the loss only to generated latent frames.
We use AdamW with a learning rate of $3\times10^{-5}$,
weight decay $10^{-3}$, 100 warm-up steps, and
linear learning-rate decay.
Training uses BF16 and FSDP on four A100 80 GB GPUs
with a global batch size of 4.
We maintain an exponential moving average (EMA)
of the model weights.

\paragrapht{Inference.}
We use the corresponding Real or Synthetic checkpoint
with EMA weights, 35 denoising steps, and a
classifier-free guidance scale of 7.0 to jointly
generate two 77-frame ego streams.

\end{document}